%% file: paper.tex
\documentclass{article}
\usepackage[utf8]{inputenc}

\usepackage[margin=1in]{geometry}
\usepackage{graphicx}
\usepackage{booktabs}
\usepackage{microtype}
\usepackage{indentfirst}
\usepackage{wrapfig}
\usepackage{makecell}
\usepackage{pdflscape}
\usepackage{arydshln}
\usepackage{longtable}
\usepackage[misc]{ifsym}

\usepackage{hyperref}
\usepackage[dvipsnames]{xcolor}
\usepackage{authblk}
\usepackage{mathtools}
\usepackage{amsmath,amsfonts,amssymb,amsthm}
\usepackage[font=small]{caption}
\usepackage[T1]{fontenc}
\usepackage{float}
\usepackage{afterpage}
\usepackage{tikz}
\usepackage{multirow}
\usepackage{array}
\usepackage{subcaption}
\usepackage{fancyhdr}
\usepackage{algorithm}
\usepackage{algorithmic}
\usepackage{subcaption}
\usepackage{bbm}
\usepackage{lineno}

\renewcommand{\headrulewidth}{0pt}
\renewcommand{\footrulewidth}{0.4pt}

\title{Multi-modal AI for comprehensive breast cancer prognostication}

\author[1]{Jan~Witowski}
\author[1]{Ken~G.~Zeng}
\author[1]{Joseph~Cappadona}
\author[2]{Jailan~Elayoubi}
\author[2]{Khalil~Choucair}
\author[3]{Elena~Diana~Chiru}
\author[4]{Nancy~Chan}
\author[5]{Young-Joon~Kang}
\author[6]{Frederick~Howard}
\author[7]{Irina~Ostrovnaya}
\author[8,9]{Carlos~Fernandez-Granda}
\author[4]{Freya~Schnabel}
\author[1]{Zoe~Steinsnyder}
\author[4]{Ugur~Ozerdem}
\author[8]{Kangning~Liu}
\author[2]{Waleed~Abdulsattar}
\author[2]{Yu~Zong}
\author[2]{Lina~Daoud}
\author[2]{Rafic~Beydoun}
\author[2]{Anas~Saad}
\author[4]{Nitya~Thakore}
\author[4]{Mohammad~Sadic}
\author[4]{Frank~Yeung}
\author[4]{Elisa~Liu}
\author[4]{Theodore~Hill}
\author[4]{Benjamin~Swett}
\author[4]{Danielle~Rigau}
\author[4]{Andrew~Clayburn}
\author[10]{Valerie~Speirs}
\author[3]{Marcus~Vetter}
\author[3]{Lina~Sojak}
\author[11]{Simone~Soysal}
\author[11]{Daniel~Baumhoer}
\author[12]{Jia-Wern~Pan}
\author[12]{Haslina~Makmur}
\author[12]{Soo-Hwang~Teo}
\author[4]{Linda~Ma~Pak}
\author[13]{Victor~Angel}
\author[14]{Dovile~Zilenaite-Petrulaitiene}
\author[14]{Arvydas~Laurinavicius}
\author[4]{Natalie~Klar}
\author[15]{Brian~D.~Piening}
\author[15]{Carlo~Bifulco}
\author[5]{Sun-Young~Jun}
\author[5]{Jae~Pak~Yi}
\author[5]{Su~Hyun~Lim}
\author[16]{Adam~Brufsky}
\author[17]{Francisco~J.~Esteva}
\author[18]{Lajos~Pusztai}
\author[8,9,19]{Yann~LeCun}
\author[1,8,4,\Letter]{Krzysztof~J.~Geras}

\affil[1]{Ataraxis AI}
\affil[2]{Karmanos Cancer Institute}
\affil[3]{Cancer Center Baselland}
\affil[4]{NYU Grossman School of Medicine}
\affil[5]{The Catholic University of Korea}
\affil[6]{UChicago Medicine}
\affil[7]{Memorial Sloan Kettering Cancer Center}
\affil[8]{NYU Center for Data Science}
\affil[9]{NYU Courant Institute of Mathematical Sciences}
\affil[10]{The University of Aberdeen}
\affil[11]{University Hospital Basel}
\affil[12]{Cancer Research Malaysia}
\affil[13]{Omica.bio}
\affil[14]{Vilnius University}
\affil[15]{Providence Health}
\affil[16]{University of Pittsburgh Medical Center}
\affil[17]{Northwell Health}
\affil[18]{Yale School of Medicine}
\affil[19]{Meta AI}

\affil[ \Letter ]{\texttt{krzysztof.geras@nyulangone.org}}

\date{}

\begin{document}

\maketitle
\thispagestyle{fancy}

\newpage
\begin{abstract}
Treatment selection in breast cancer is guided by molecular subtypes and clinical characteristics. However, current tools including genomic assays lack the accuracy required for optimal clinical decision-making. We developed a novel artificial intelligence (AI)-based approach that integrates digital pathology images with clinical data, providing a more robust and effective method for predicting the risk of cancer recurrence in breast cancer patients. Specifically, we utilized a vision transformer pan-cancer foundation model trained with self-supervised learning to extract features from digitized H\&E-stained slides. These features were integrated with clinical data to form a multi-modal AI test predicting cancer recurrence and death.
The test was developed and evaluated using data from a total of 8,161 female breast cancer patients across 15 cohorts originating from seven countries. Of these, 3,502 patients from five cohorts were used exclusively for evaluation, while the remaining patients were used for training.
Our test accurately predicted our primary endpoint, disease-free interval, in the five evaluation cohorts (C-index: 0.71 [0.68-0.75], HR: 3.63 [3.02-4.37, p<0.001]). In a direct comparison (n=858), the AI test was more accurate than Oncotype DX, the standard-of-care 21-gene assay, achieving a C-index of 0.67 [0.61-0.74] versus 0.61 [0.49-0.73], respectively. Additionally, the AI test added independent prognostic information to Oncotype DX in a multivariate analysis (HR: 3.11 [1.91-5.09, p<0.001)]). The test demonstrated robust accuracy across major molecular breast cancer subtypes, including TNBC (C-index: 0.71 [0.62-0.81], HR: 3.81 [2.35-6.17, p=0.02]), where no diagnostic tools are currently recommended by clinical guidelines. These results suggest that our AI test improves upon the accuracy of existing prognostic tests, while being applicable to a wider range of patients.
\end{abstract}

\begin{figure}[!t]
    \centering
    \includegraphics[width=0.9475\linewidth]{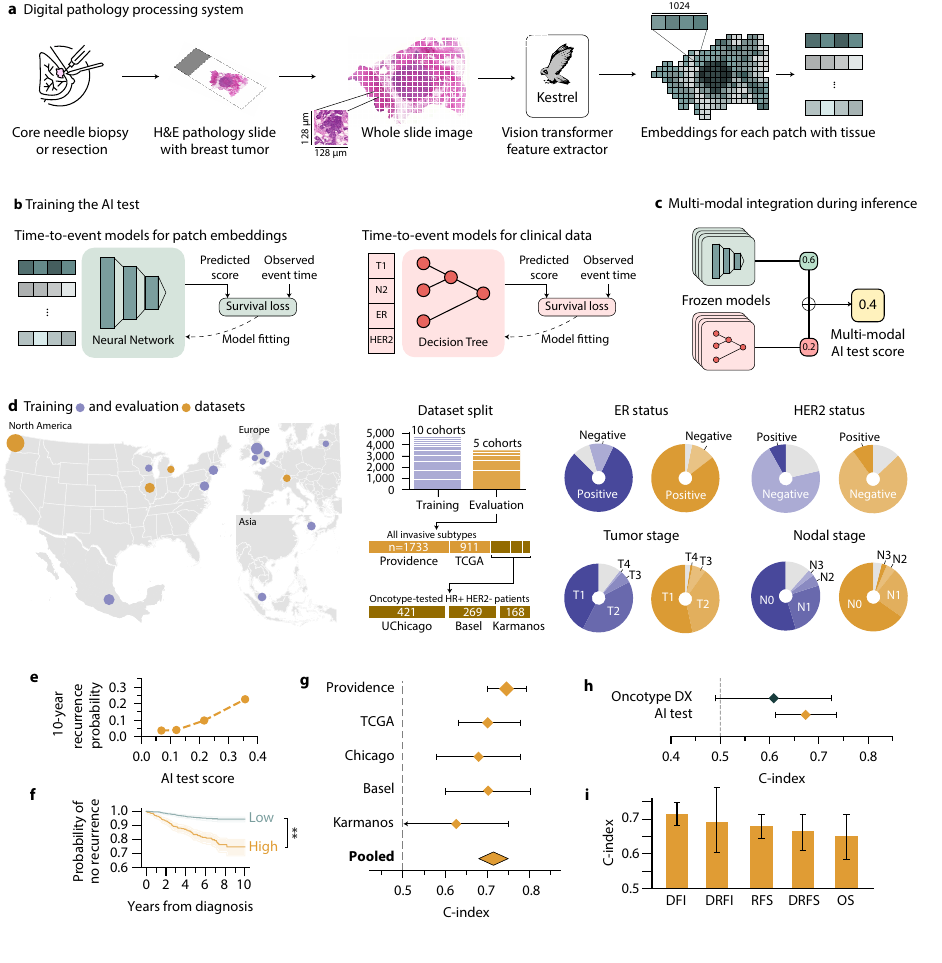}
    \caption{\textbf{We present a multi-modal AI test for invasive breast cancer.} \textbf{a}, The key component of the test is a system which processes high-resolution digital images of breast cancer specimens. Features are extracted from the digitized slides using \emph{Kestrel}, a foundation model trained using self-supervised learning on a pan-cancer dataset of 400 million pathology image patches. \textbf{b}, Extracted pathology features and clinical features are used to train supervised time-to-event models predicting breast cancer recurrence or death. \textbf{c}, \emph{The AI test} produces a multi-modal risk score, integrating pathological and clinical risk scores. \textbf{d}, We developed the AI test using 4,659 patients across 10 cohorts from six countries, and evaluated it on 3,502 patients from five patient cohorts which were not used during training. Evaluation sets consisted of two cohorts (Providence and TCGA) with all invasive breast cancer subtypes, and three cohorts (UChicago, Basel, and Karmanos) with only HR+ HER2\textminus\ patients tested with Oncotype~DX. \textbf{e}, The 10 year recurrence probability increases monotonically as the AI test score increases. \textbf{f}, Patients with predicted high risk had significantly worse outcomes compared to patients with predicted low risk. \textbf{g}, The AI test achieved strong prognostic results across all evaluation datasets. \textbf{h}, In a direct comparison (n=858) to a standard-of-care genomic assay, Oncotype~DX, the AI test was a better predictor of cancer recurrence. \textbf{i}, The AI test performs consistently across various endpoints (DFI - disease-free interval, DRFI - distant recurrence-free interval, RFS - recurrence-free survival, DRFS - distant recurrence-free survival, OS overall survival). Endpoint definitions are in Appendix~\ref{appendix:endpoints}.}
    \label{fig:overview}
\end{figure}

\section{Main}
Over the past few decades, breast cancer treatment has evolved significantly with the introduction of various chemo, endocrine, and targeted therapies. Treatment selection is primarily informed by clinical guidelines that rely on clinicopathological factors such as tumor size, nodal status, and estrogen and HER2 receptor status. While this approach optimizes outcomes at the population level to a certain degree, it overlooks critical, actionable information for individual patients, exposing a significant gap in personalized care.

In an attempt to address this limitation, genomic assays were developed in the early 2000s, offering a more refined approach to patient risk stratification. Tools such as Oncotype~DX~\cite{paik2004multigene}, Mammaprint~\cite{mook2007individualization} and Prosigna~\cite{wallden2015development} assess the risk of distant recurrence based on gene expression data. However, these assays were developed primarily for hormone receptor-positive (HR+) breast cancer patients, mainly to de-escalate adjuvant chemotherapy. In addition to being limited to certain cancer subtypes, their accuracy is modest, often on par with simple models based on clinical characteristics~\cite{sparano2020Clinical, use2017esserman}. These tests also require the processing of physical tissue specimens, which creates additional work for pathology departments and depletes tissue that could be used in the future for advanced molecular profiling.

Complementary to genomic assays, histopathological features are routinely evaluated by pathologists to stratify patient risk. These features include histologic grade, tumor infiltrating lymphocytes level, mitotic count, and Ki67 expression. These primarily morphological features have been shown to add independent prognostic information to genomic scores, and recent prognostic models integrate both~\cite{pusztai2024development, miglietta2023242mo}. However, these features are limited in their scope and prognostic ability, which drives the need for stronger and more informative prognostic features. In recent years, advances in AI, particularly in self-supervised learning, have enabled the development of more effective methods for learning meaningful features from imaging data~\cite{chen2020simple, he2020momentum, grill2020bootstrap, he2022masked, caron2021emerging, oquab2024dino}. These features are not based on pre-specified characteristics and extracting them does not require manual annotation by pathologists. Instead, AI models autonomously determine the most salient features by learning from millions of images. Early evidence shows that these AI-enabled features extracted from digital pathology images can be used to predict key patient characteristics and long-term outcomes~\cite{vorontsov2024foundation, chen2024towards,wang2024pathology}.

In response to the gaps in today's treatment selection tools, we developed and evaluated an AI test for predicting the risk of cancer recurrence in invasive breast cancer patients. It is built upon the latest advances in self-supervised learning and digital pathology. Specifically, our test extracts strongly generalizable morphological features from standard H\&E-stained slides of core needle biopsy or surgical specimens utilizing \textit{Kestrel}~\cite{cappadona2024squeezing}, a state-of-the-art pan-cancer AI foundation model. Kestrel is a variant of vision transformer~\cite{dosovitskiy2021an} trained with the self-supervised learning method DINOv2~\cite{oquab2024dino} using 400M digital pathology image patches. The image patches used to train Kestrel were sourced from a large, heterogeneous database of pan-cancer whole slide images, including breast cancer images. Additionally, we tuned Kestrel's hyperparameters using a novel methodology that evaluates model performance on a suite of digital pathology benchmarks. This approach iteratively scales model and dataset size, yielding a model that is specially tuned for downstream tasks in digital pathology.

In this study, we present a multi-modal AI test that integrates pathology-based features, extracted by Kestrel, with routine clinical characteristics, namely cancer T and N stage, patient age, ER, PR and HER2 status and presence of ductal or lobular histology (Figure~\ref{fig:overview}). The test produces a continuous score between 0 and 1 that quantifies the risk of cancer recurrence. The test was developed and evaluated using data from a total of 8,161 patients across 15 cohorts originating from seven countries. Of these, 3,502 patients from five cohorts were used exclusively for external evaluation, while the remaining patients were used for model training. Detailed patient characteristics are described in Table~\ref{tab:cohorts_breakdown}.

We evaluate our test's prognostic capabilities in a variety of clinically meaningful patient groups. We also compare our test's accuracy to a standard-of-care genomic assay, showing how the introduced multi-modal AI test enhances breast cancer prognostication.

\section{Results}

\subsection{Breast cancer cohorts}

To execute this project we assembled a diverse collection of both public datasets (TCGA-BRCA~\cite{cancer2012comprehensive, ciriello2015comprehensive}, METABRIC~\cite{curtis2012genomic}, BASIS~\cite{nikzainal2016landscape}, Providence ~\cite{bifulco2021identifying}) and private datasets from several sources (NYU Langone Health, The Catholic University of Korea, Gundersen Health, National Center of Pathology in Lithuania, Breast Cancer Now Biobank, Cancer Research Malaysia, Omica.bio Cancer Atlas, Wales Cancer Biobank, Karmanos Cancer Institute, University Hospital Basel, and UChicago Medicine). Altogether, we collected 5162 slides from 4659 breast cancer patients across 10 datasets for training, and 3632 slides from 3502 breast cancer patients from 5 datasets for evaluation, making this one of the largest data collection efforts for predicting breast cancer recurrence. All patients in data are women. Two of the evaluation datasets, Providence (n=1733) and TCGA-BRCA (n=911), represent a broad spectrum of invasive breast cancer patients. The remaining three test cohorts --- Karmanos (n=168), Basel (n=269), and UChicago (n=421) --- included only HR+ HER2\textminus\ patients who were previously tested with Oncotype~DX. See Table~\ref{tab:cohorts_breakdown} for more detailed statistics of the training and evaluation cohorts.

\subsection{Study design and results}

Unless explicitly stated otherwise, the presented results are from the multi-modal AI test. We first report its prognostic capability, i.e., whether our model accurately predicts the risk of cancer recurrence and distinguishes between high- and low-risk patients. Then, we compare its accuracy against a standard-of-care genomic risk signature in various molecular, histological and clinical subgroups. Finally, we discuss potential clinical use cases. When reporting performance across multiple datasets, we present pooled results using a random effects model. The additional results for specific data sets are in Figure \ref{fig:mixed-effect-subgroups-dfi-cindex} and Figure \ref{fig:mixed-effect-subgroups-dfi-ai}. 

To evaluate how well the test orders patients according to their risk, we report the concordance index (C-index). Additionally, we estimate the hazard ratio (HR) for the AI test in univariate and multivariate Cox proportional hazards models. While the AI test takes values between 0 and 1, the HR presented is per every 0.2 unit increase. Unless explicitly stated otherwise, we use the continuous AI test risk score, rather than its dichotomized version. To compare hazard ratio with Oncotype DX, we dichotimize the score from the AI test. We defined high- and low-risk groups using the 80\textsuperscript{th} percentile cutoff in the HR+ HER2- subpopulation within our largest internal training dataset (NYU). We refer to Methods (Section~\ref{sec:methods}) for a detailed explanation of the performance metrics used in this paper.

\subsection{The AI test is predictive of breast cancer recurrence and survival}

The AI test was found to be prognostic in all five patient cohorts used as external evaluation sets. For the primary endpoint, disease-free interval (DFI), the risk score generated by our model achieved a pooled C-index of 0.71 [0.68-0.75] and a pooled HR of 3.63 [3.02-4.37, p<0.001].

Two cohorts --- Providence and TCGA-BRCA --- were composed of broad population of invasive breast cancer patients. Our AI test achieved a C-index of 0.74 [0.70-0.79] and a HR 4.02 [3.09-5.23, p<0.001] in the Providence cohort, and a C-index of 0.70 [0.63-0.77] and a HR of 3.00 [2.10-4.28, p<0.001] in the TCGA-BRCA cohort. 

Three cohorts --- Karmanos, Basel, and UChicago --- included only HR+ HER2\textminus\ patients who were previously tested with Oncotype~DX. The AI test achieved a C-index of 0.62 [0.49-0.75] and a HR of 3.82 [1.33-10.98, p=0.013] in the Karmanos cohort (n=168), a C-index of 0.70 [0.60-0.80] and a HR of 3.98 [1.92-8.25, p<0.001] in the Basel cohort (n=269), and a C-index of 0.67 [0.58-0.77] and a HR of 3.26 [1.45-7.31, p=0.004] in the UChicago cohort (n=421).

The AI test was also prognostic for secondary endpoints (as defined in Appendix~\ref{appendix:endpoints}). For distant-disease free interval (DRFI), the test achieved a C-index of 0.70 [0.61-0.78] and a HR of 4.02 [2.64-6.13, p=0.001]. For overall survival (OS), the test achieved a C-index of 0.65 [0.58-0.72] and a HR of 2.52 [1.88-3.38, p=0.001]. For recurrence-free survival (RFS), the test achieved a C-index of 0.68 [0.65-0.71] and a HR of 2.65 [2.33-3.01, p<0.001]. For distant recurrence-free survival (DRFS), the test achieved a C-index of 0.66 [0.61-0.71] and a HR of 2.64 [2.14-3.26, p<0.001]. Forest plots for all endpoints and metrics are reported in Figure \ref{fig:mixed-effect-subgroups-all-endpoints-cindex} and Figure \ref{fig:mixed-effect-subgroups-all-endpoints-hr}.

\subsection{The AI test is more accurate than a genomic assay in predicting cancer recurrence in HR+ patients}

\begin{figure}[ht!]
    \centering
    \includegraphics{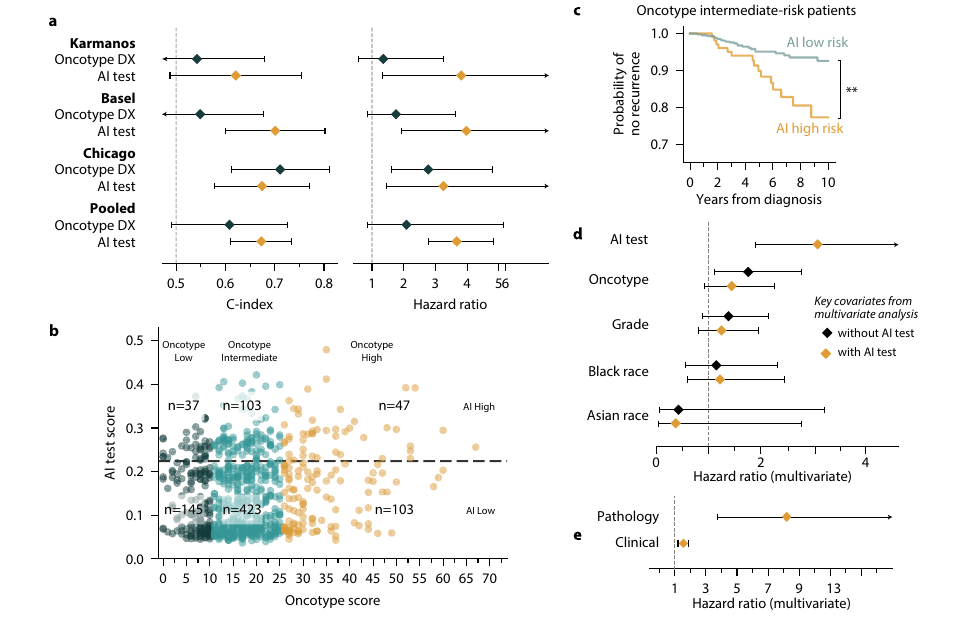}
    \caption{\textbf{The AI test enhances the prognostic accuracy over standard-of-care genomic assays for predicting cancer recurrence.} Oncotype~DX, a standard-of-care 21-gene assay, classifies patients into low-, intermediate-, and high-risk groups. The AI test demonstrated statistically significant discrimination between high- and low-risk patients without the need to introduce an intermediate-risk group, thereby enhancing clarity in decision-making. We analyze the differences in classification and patient outcomes between the two tests using data from the Karmanos, Basel, and UChicago cohorts. 
    \textbf{a}, Comparison of prognostic ability between Oncotype~DX and the AI test in univariate models. We compare the hazard ratio for every 0.2 increase in our score, compared to a 20 point increase for Oncotype DX. \textbf{b}, A scatter plot illustrating risk scores for Oncotype~DX-tested patients. Each point represents one patient. The majority of patients with intermediate Oncotype~DX scores were reclassified into the low-risk group by the AI test. \textbf{c}, For intermediate-risk Oncotype~DX patients, the AI test was able to accurately distinguish between low- and high-risk patients (HR 2.84 [1.47-5.49, p=0.002]) \textbf{d}, Hazard ratios associated with common clinical covariates in a multivariate Cox analysis, with and without including the AI test. The AI test is significantly associated with DFI after adjusting for Oncotype~DX score, grade (based on the Nottingham grading system, categorized from 1 to 3) and race in a multivariate Cox regression model. \textbf{e}, Comparing the AI test's modalities in a multivariate Cox model shows that the pathology score was more informative than the clinical score within the Oncotype~DX cohort (see Table \ref{tab:cox-path-clin}).}
  \label{fig:oncotype_comparison}
\end{figure}

We compared the prognostic performance and potential clinical utility of our model to Oncotype~DX, a standard-of-care 21-gene assay developed to predict distant recurrence risk. Three external cohorts (Karmanos, Basel, UChicago) contain a total of 858 patients who were tested with Oncotype~DX Recurrence Score assays performed as part of routine care.

In the Karmanos cohort, the AI test's C-index for DFI was 0.62 [0.49-0.76] and the HR was 3.82 [1.33-10.98, p=0.013], compared with Oncotype~DX's C-index of 0.54 [0.41-0.68] and HR of 1.36 [0.56-3.27, p=0.500]. 
In the Basel cohort, the AI test's C-index was 0.70 [0.60-0.80] and the HR was 3.98 [1.92-8.25, p<0.001], compared with Oncotype~DX's C-index of 0.55 [0.42-0.68] and HR of 1.76 [0.85-3.64, p=0.13].
In the UChicago cohort, the AI test's C-index was 0.67 [0.58-0.77] and the HR was 3.26 [1.45-7.31, p=0.004], compared with Oncotype~DX's C-index of 0.71 [0.61-0.81] and HR of 2.78 [1.61-4.81, p<0.001].
Together, the AI test achieved a pooled C-index of 0.67 [0.61-0.74] and HR of 3.67 [2.79-4.84, p=0.002], compared to Oncotype~DX's C-index of 0.608 [0.491-0.725] and HR of 2.09 [0.85-5.15, p=0.21]. These comparisons are illustrated in Figure~\ref{fig:oncotype_comparison}A and further extended in Figure \ref{fig:mixed-effect-subgroups-dfi-oncotype}.

Oncotype~DX has been shown to predict the benefit of adjuvant chemotherapy in HR+ patients. Patients with high Oncotype~DX scores had significant chemotherapy benefits, while patients with low Oncotype~DX scores did not~\cite{albain2010prognostic}. Large randomized clinical trials, TAILORx~\cite{sparano2018adjuvant} and RxPONDER~\cite{kalinsky202121}, have shown that some patients with intermediate Oncotype~DX scores might also safely avoid adjuvant chemotherapy. For patients tested with Oncotype~DX, the AI model would re-classify 666 out of 858 (77.6\%) patients into different risk categories (Figure~\ref{fig:oncotype_comparison}B). All intermediate Oncotype-risk patients (61.3\% of all Oncotype~DX patients) would be reclassified as either low- or high-risk by our AI test. 423 out of 526 (80.4\%) intermediate Oncotype-risk patients would be classified as low-risk patients and 103 of 526 (19.6\%) as high-risk patients using the AI test. The outcomes for high/low risk patients are illustrated in Figure~\ref{fig:oncotype_comparison}C. Within this intermediate Oncotype-risk group, the continuous AI test score was significantly associated with recurrence (HR: 3.45 [1.85-6.42, p<0.001]) in the Karmanos, Basel and UChicago cohorts after adjusting for the dataset (see Table~\ref{tab:cox-multimodal}).

\begin{figure}[t!]
    \centering
    \includegraphics{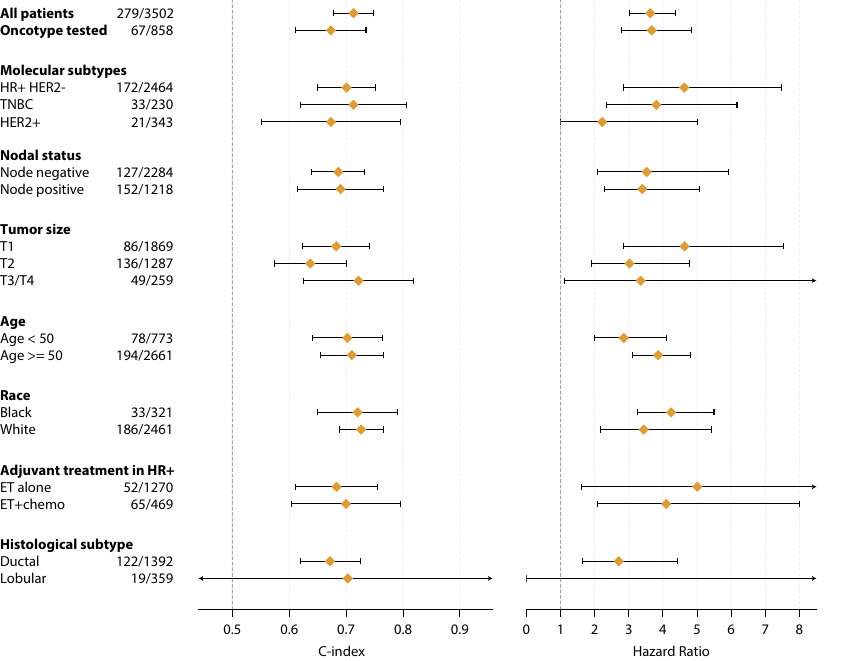}
    \caption{\textbf{The AI test performs well in all major clinically relevant groups.} Forest plots display our AI test's performance across clinical, molecular, and demographic groups. Next to each group name we report the number of disease-free interval events and total number of patients. Subgroup performances were pooled across all evaluation cohorts using a random effects model. For subgroup performance based on histological subtype, we excluded patients from the Providence cohort because histological subtype data were unavailable.}
    \label{fig:subgroup_performance}
\end{figure}

Finally, we demonstrate that the AI test provides independent prognostic value from Oncotype DX, as well as from other common clinical indicators. To do this, we incorporated the AI test score into a multivariate Cox proportional hazards model along with Oncotype~DX score, Nottingham cancer grade, dataset and race (see Section~\ref{sec:methods} for more details). After adjusting for these factors, the AI test has a HR of 3.11 [1.91-5.09, p<0.001]. In comparison, Oncotype~DX has a HR of 1.47 [0.93-2.30, p=0.10], which is not statistically significant. Race groups or grade were not significant. Key findings are presented in Figure~\ref{fig:oncotype_comparison}D. The full results are in Table~\ref{tab:cox-oncotype-multimodal} in Appendix~\ref{appendix:cox-multivariate-analysis}. To further demonstrate the the AI test is not solely dependent on clinical covariates, we performed multivariate analysis with the clinical and pathology score within the AI test. As shown in Figure~\ref{fig:oncotype_comparison}E, the pathology score is significant after adjusting for the clinical score. This shows that the pathology adds prognostic information that is independent from the information in clinical covariates used in our AI test. 

\subsection{The AI test performs well in various clinically meaningful subgroups: potential clinical utility}

\begin{figure}[ht!]
    \centering
    \includegraphics{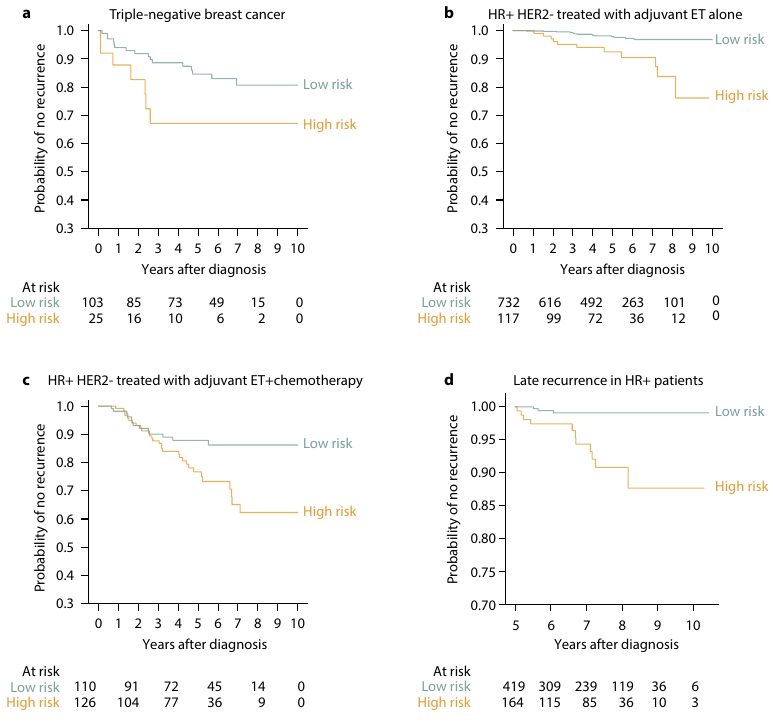}
    \caption{\textbf{The AI test is prognostic in clinically meaningful subgroups.} All plots in this figure are for the Providence cohort (the largest evaluation cohort). \textbf{a}, In triple-negative breast cancer, there is an ongoing discussion about potential strategies to de-escalate the KEYNOTE-522 regimen, for example with shorter or anthracycline-free chemotherapy regimens or complete ommitance of adjuvant treatment. While some biomarkers, such as tumor-infiltrating lymphocytes, are associated with long-term outcomes and pathological complete response, there is still a need for a more robust assessment of systemic therapy benefits. \textbf{b-d}, In hormone receptor-positive patients, questions about treatment selection involve the addition of adjuvant chemotherapy, extended endocrine therapy (ET), and new agents, such as CDK4/6 inhibitors. We hypothesize that high-risk HR+ patients who received adjuvant ET alone (\textbf{b}) might benefit from the addition of chemotherapy, and high-risk HR+ patients who received chemoendocrine therapy (\textbf{c}) might benefit from the addition of CDK4/6 inhibitors. Finally, HR+ patients who received five years of adjuvant endocrine therapy (\textbf{d}) are candidates for extended endocrine treatment. A few assays have been shown to be associated with late recurrence and predict the benefit of extended endocrine therapy~\cite{bartlett2019breast}.} 
    \label{fig:treatment-outcomes}
\end{figure}

\begin{figure}[t]
    \centering
    \includegraphics{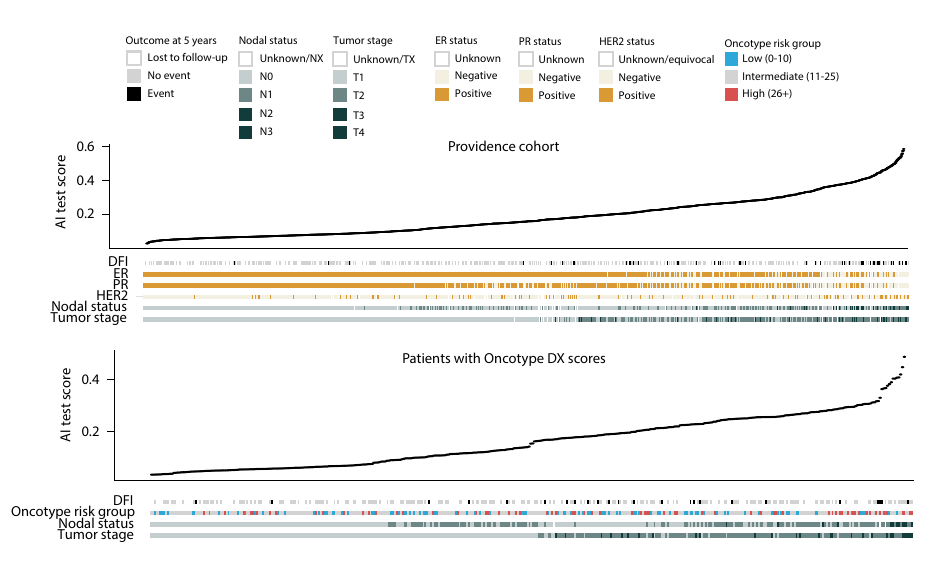}
    \caption{\textbf{Relationship between predicted risk of recurrence and established prognostic factors.} The top plot analyzes the Providence cohort to display the relationship between the AI test score and the established factors such as ER/HER2 status and staging. The bottom plot combines the Basel, Karmanos and Chicago cohorts and illustrates the Oncotype score in relationship with the AI test score. Higher scores are observed in patients with HR\textminus\ and HER2+ status and in patients with more advanced T or N staging. Higher AI test scores do not appear to be strongly associated with higher Oncotype DX risk, as indicated by the near-zero correlation ($R^2=0.02$) between our AI test scores and the Oncotype DX scores.}
    \label{fig:heat-map}
\end{figure}

Traditional genomic assays are typically developed for specific patient subgroups, such as Oncotype DX for HR+ HER2\textminus\ patients, and cannot be used in patients with less common molecular, histological, or clinical characteristics. Given that the AI test was developed using data from non-metastatic breast cancer patients without additional exclusion criteria, we can evaluate its prognostic accuracy in various subgroups. This includes patients with varying age and menopausal status, nodal status, tumor size, estrogen and HER2 receptor status, race, and administered adjuvant therapy (Figure~\ref{fig:subgroup_performance}). Furthermore, our analysis indicates that the AI test score is associated with known clinical factors that influence cancer prognosis, including higher scores in ER\textminus\ and PR\textminus\ patients, as well as patients with more advanced staging (both N and T) (Figure~\ref{fig:heat-map}).

Currently, there is no prognostic/predictive tool supported by NCCN guidelines for use in triple-negative and HER2+ breast cancer patients~\cite{waks2023assessment}. The AI test reliably predicted DFI in TNBC (n=230, C-index: 0.71 [0.62-0.81], HR: 3.81 [2.35-6.17, p=0.02]) and HER2+ patients (n=343, C-index: 0.67 [0.55-0.80], HR: 2.22 [0.99-5.01, p=0.05]). 

NCCN guidelines also do not currently recommend any tests for the selection of adjuvant systemic therapy in most premenopausal HR+ patients (intermediate-risk node-negative and low- or intermediate-risk 1-3 LN+). The AI test reliably predicted DFI in HR+ premenopausal patients (if the menopausal status is unknown, we consider patients younger than 50 years as premenopausal), with a C-index of 0.655 [0.583-0.726] and a HR of 2.87 [1.81-4.55, p=0.003] (n=677).

Furthermore, genomic assays have been reported to perform poorly in Black patients~\cite{flanagin2021updated, satpathy2023comparison}. As shown in Figure~\ref{fig:subgroup_performance}, the AI test reliably predicted DFI in Black patients (n=299), with a C-index of 0.720 [0.65-0.79] and a HR 4.24 [3.27-5.50, p=0.002]. Among White patients (n=2,461), the AI test achieved a C-index of 0.73 [0.69-0.79] and a HR of 3.44 [2.18-5.42, p=0.003].

Some prognostic genomic assays, including Oncotype DX, have also been shown to be predictive of adjuvant chemotherapy benefit~\cite{albain2010prognostic, paik2006gene} in HR+ patients. We hypothesize that our test can, analogously, identify patients who might benefit from adjuvant chemotherapy, CDK4/6 inhibitors, and extended endocrine therapy. 

First, the AI test was prognostic in HR+ HER2\textminus\ patients who received adjuvant endocrine therapy (ET) alone (n=1171, C-index: 0.68 [0.61-0.76], HR: 5.01 [1.61-15.58, p=0.03]). High-risk patients in the ET group are likely to benefit from the addition of adjuvant chemotherapy. 

Second, the AI test was prognostic in HR+ HER2\textminus\ patients who received adjuvant chemoendocrine therapy (n=469, C-index: 0.70 [0.60-0.80], HR: 4.10 [2.09-8.02,p=0.007]). Patients who still have ``residual'' high risk while receiving both endocrine and chemotherapy are likely to benefit from the addition of CDK4/6 inhibitors.

Finally, the AI test was prognostic for late recurrence in HR+ HER2\textminus\ patients (n=1158, C-index 0.79 [0.71-0.86], HR 8.45 [3.11-22.93, p<0.001]), suggesting that patients with higher risk of cancer recurrence after 5 years might benefit from extended endocrine therapy. 

To illustrate the potential utility of our test in clinical decision-making, we defined high- and low-risk groups using the 80th percentile cutoff from our largest internal training dataset (NYU). For the analyses regarding treatment within HR+ HER2- subgroup, we used the 80th percentile within the HR+ HER2- subpopulation, while for the triple-negative breast cancer (TNBC) subgroup, we also applied the 80th percentile cutoff from its corresponding population. We selected this threshold as it was yielding statistically significant hazard ratios in our training set. Additionally, this threshold balanced the distribution of high- and low-risk groups across all subgroups, consistently demonstrating worse outcomes for high-risk patients compared to low-risk patients. As demonstrated in Figure \ref{fig:treatment-outcomes}, when we apply this cutoff to the Providence cohort, we see a significant difference in outcomes between the low- and high-risk groups.

\section{Discussion}

In this study, we present a novel prognostic multi-modal AI test for invasive breast cancer. The overall performance of our test is robust across breast cancer patients. We also compared the AI test to Oncotype DX in three cohorts where the Oncotype score is available. The test outperformed Oncotype DX in two cohorts, Karmanos and Basel, and performed as well as Oncotype DX in one additional cohort, UChicago. Notably, the presented test is one of the few prognostic tools that have been clinically validated in triple-negative and HER2+ breast cancer patients. This capability, combined with strong performance regardless of nodal status and patient age, supports the viability of the AI test as a single tool to inform treatment decisions in all breast cancer patients.

The proposed AI test is designed to predict cancer recurrence risk, similar to genomic assays such as Oncotype~DX and MammaPrint. However, just like these assays, it is not explicitly trained to model treatment effects. We hypothesize that when validated using randomized prospective data, our prognostic test will show predictive capabilities in the same manner that the genomic assays do. That is, high-risk patients will benefit from more aggressive treatment options (such as adjuvant chemotherapy), while low-risk patients will have no benefit. While this study uses observational data and did not evaluate the test's predictive capabilities (i.e., its ability to determine whether a patient is likely to benefit from a specific treatment), we present several potential clinical implications of using such a test in a clinical setting.

The AI test has the potential to refine clinical decision-making in several key areas. In HR+ patients who are candidates for adjuvant chemotherapy, the AI test was more accurate than Oncotype DX. This can potentially improve the selection of patients who can benefit from chemotherapy in addition to endocrine therapy. In a multivariate analysis, the AI test was shown to add independent prognostic information with respect to Oncotype and cancer grade, which indicates that it captures information that is strongly predictive and independent from these factors. Importantly, the AI test was prognostic in premenopausal HR+ patients, expanding patient groups who could be evaluated for cancer recurrence risk. Finally, we show that the AI test is prognostic in HR+ patients who received adjuvant chemoendocrine therapy, and identifies patients with high ``residual'' recurrence risk. We hypothesize that these patients might benefit from adjuvant CDK4/6 inhibitors. This use case is especially meaningful given recent FDA approval for ribociclib, which can be used in a wide group of HR+ patients, including node-negative cases. Identifying high-risk patients beyond basic clinical characteristics will be important in selecting patients who would benefit from CDK4/6 inhibitor treatment.

Besides HR+ patients, the test was prognostic in triple-negative and HER2+ patients, for which there are no NCCN guideline-supported tests for treatment selection. Current standard of care treatment in triple-negative breast cancer involves intense immunochemotherapy (KEYNOTE-522 regimen~\cite{takahashi2023permbrolizumab, schmid2022event}). There are multiple ongoing clinical trials working on treatment de-escalation through avoiding adjuvant pembrolizumab in patients with pathological complete response (OptimICE-pCR~\cite{santa-maria2023optimizing, tolaney2024optimice}) or using shorter, anthracycline-free neoadjuvant chemotherapy regimen in patients with immune-enriched TNBC (NeoTRACT)~\cite{stecklein2024neotract}.

Our test stands apart from existing genomic tests as it leverages pathology data and employs self-supervised learning. Previous generation of feature extractors for digital pathology relied on pre-specified ``pathomic'' features~\cite{fernandez2022development}. Self-supervised learning paradigm does not require any pre-specification of features; rather, it enables the model to learn the most salient features. Spratt et al. successfully applied this technique to prostate cancer, with their model validated through several randomized clinical trials~\cite{spratt2023artificial, spratt2022ai}. Similarly, Garberis et al. developed an AI test for breast cancer using contrastive learning, yielding strong results~\cite{garberis2022deep}, though their study was limited to a single ER+ HER2\textminus\ cohort. Volinsky-Fremond et al. extended the use of self-supervised deep learning to endometrial tumor slides, outperforming conventional combined pathological and molecular analyses and demonstrating predictive utility for chemotherapy benefit in the PORTEC-3 trial~\cite{VolinskyFremond2024}. Our test is developed using Kestrel, an AI foundation model for digital pathology trained using self-supervised learning to extract morphological features from digitized pathology slides. As Kestrel was developed using a pan-cancer dataset, it showed promising results for a wide range of tasks across different clinical indications \cite{cappadona2024squeezing}.

Integration of digital pathology-based tests into clinical workflow offers several advantages over genomic and other molecular prognostic and predictive tests. First, the turnaround time is significantly shorter, potentially allowing for more timely treatment decisions. Second, by utilizing standard H\&E-stained slides and not requiring specialized wet lab tissue processing, the test is likely to be substantially cheaper than genomic assays, improving access to treatment selection testing and allowing for potentially pro-bono use in resource-limited settings. Third, our test can be performed on core biopsy slides, which allows for earlier risk assessment and might inform neoadjuvant therapy decisions. Finally, this digital approach preserves valuable tumor samples, which might be needed for deep sequencing in cases of disease progression.

In summary, this study underscores the potential of a digital pathology-based AI test in breast cancer prognosis. By delivering accurate, accessible, and cost-effective risk stratification in diverse patient populations, the test bridges significant gaps in current clinical workflows. With more rigorous validation through clinical trials, it could become an essential tool in personalized cancer care.

\newpage
\section{Methods}
\label{sec:methods}

\subsection{Data}

The breakdown of patient demographic, molecular, and clinical characteristics of the training and test datasets is described in Table~\ref{tab:cohorts_breakdown}.

\begin{table}[htbp]
    \centering
    \footnotesize
    \caption{Datasets used for training and evaluation. All values are N (\%) unless indicated otherwise. Follow-up time statistics are computed taking into account all patients (censored and not censored).}
    \label{tab:cohorts_breakdown}
\begin{tabular}{lrrr}
\toprule
 & Category & \textbf{Training cohorts} (N=4,659) & \textbf{Evaluation cohorts}(N=3,502) \\
\hline
 \multicolumn{2}{l}{\textbf{Age} \textit{median, [IQR]}} & 58 [49-68] & 60 [51-68] \\
\hline
\multicolumn{2}{l}{\textbf{Follow-up time} \textit{median, [IQR]}} & 5.11 [3.62-8.31] & 4.51 [2.42-7.15] \\

\hline
\multicolumn{4}{l}{\textbf{ER receptor status}} \\
& Positive & 3748 (80.45\%) & 2992 (85.44\%) \\
& Negative & 540 (11.59\%) & 399 (11.39\%) \\
& Unknown & 371 (7.96\%) & 111 (3.17\%) \\
\hline
\multicolumn{4}{l}{\textbf{PR receptor status}} \\
& Positive & 2676 (57.44\%) & 2648 (75.61\%) \\
& Negative & 1010 (21.68\%) & 728 (20.79\%) \\
& Unknown & 973 (20.88\%) & 126 (3.60\%) \\
\hline
\multicolumn{4}{l}{\textbf{HER2 receptor status}} \\
& Positive & 377 (8.09\%) & 348 (9.94\%) \\
& Negative & 3286 (70.53\%) & 2702 (77.16\%) \\
& Equivocal or Unknown & 996 (21.38\%) & 452 (12.91\%) \\
\hline
\multicolumn{4}{l}{\textbf{T stage}} \\
& T1 & 1880 (40.35\%) & 1869 (53.37\%) \\
& T2 & 1780 (38.21\%) & 1287 (36.75\%) \\
& T3 & 252 (5.41\%) & 225 (6.42\%) \\
& T4 & 51 (1.09\%) & 40 (1.14\%) \\
& Unknown/TX & 464 (9.96\%) & 81 (2.31\%) \\
\hline
\multicolumn{4}{l}{\textbf{N stage}} \\
& N0 & 2550 (54.73\%) & 2284 (65.22\%) \\
& N1 & 1174 (25.20\%) & 839 (23.96\%) \\
& N2 & 251 (5.39\%) & 156 (4.45\%) \\
& N3 & 141 (3.03\%) & 86 (2.46\%) \\
& Unknown/NX & 543 (11.65\%) & 137 (3.91\%) \\
\hline
\multicolumn{4}{l}{\textbf{Ductal histology}} \\
& Yes & 3567 (76.56\%) & 1392 (39.75\%) \\
& No & 904 (19.40\%) & 377 (10.77\%) \\
& Unknown & 188 (4.04\%) & 1733 (49.49\%) \\
\hline
\multicolumn{4}{l}{\textbf{Lobular histology}} \\
& Yes & 600 (12.88\%) & 359 (10.25\%) \\
& No & 3871 (83.09\%) & 1410 (40.26\%) \\
& Unknown & 188 (4.04\%) & 1733 (49.49\%) \\
\hline
\multicolumn{4}{l}{\textbf{Any recurrence}} \\
& Yes & 726 (15.58\%) & 279 (7.97\%) \\
& No & 3933 (84.42\%) & 3223 (92.03\%) \\
\hline
\multicolumn{4}{l}{\textbf{Distant recurrence}} \\
& Yes & 457 (9.81\%) & 174 (4.97\%) \\
& No & 4098 (87.96\%) & 3328 (95.03\%) \\
& Unknown & 104 (2.23\%) & 0 (0.00\%) \\
\hline
\multicolumn{4}{l}{\textbf{Death}} \\
& Yes & 935 (20.07\%) & 303 (8.65\%) \\
& No & 3710 (79.63\%) & 3199 (91.35\%) \\
& Unknown & 14 (0.30\%) & 0 (0.00\%) \\
\hline
\multicolumn{4}{l}{\textbf{Adjuvant chemo- and endocrine therapy}} \\
& Chemoendocrine therapy  & 1124 (24.13\%) & 768 (21.93\%) \\
& Chemotherapy only & 564 (12.11\%) & 500 (14.28\%) \\
& Endocrine therapy only & 1551 (33.29\%) & 1442 (41.18\%) \\
& Unknown & 729 (15.65\%) & 368 (10.51\%) \\
\bottomrule
\end{tabular}
\end{table}

\subsection{Task and evaluation}

Our primary task was to accurately predict the risk of cancer-related events. To accomplish this, we trained time-to-event models using clinical variables and digital pathology slides. For both modalities, we trained several models and ensembled their predictions, as detailed in Section~\ref{sec:ensembling}. We then created a multi-modal model by averaging predictions of the two ensembles. We trained and evaluated the models using the endpoint definitions in Table~\ref{tab:endpoint_definitions}.

\subsubsection{Performance metrics}

The two primary metrics we use are the concordance index (C-index) and hazard ratio (HR). Apart from computing these metrics on each individual dataset, we use a random effects model to obtain \emph{pooled} metrics which summarize performance over multiple datasets, as explained in Appendix \ref{sec:random_effects}. In the following paragraphs, we justify the use of these metrics in our application.

\paragraph{C-index.} The C-index measures how well the predicted risk ranking of patients aligns with the actual order in which they experienced events or are censored. A C-index of 1.0 indicates that patients are perfectly ordered according to predicted risk, whereas a C-index of 0.5 indicates that the model's ordering is no better than random. The advantage of the C-index is that it is easy to interpret, as it is analogous to AUROC in how it is computed, while accommodating censoring. A detailed explanation of how the C-index is computed is in Appendix~\ref{sec:c-index}.

\paragraph{Hazard ratio of a continuous score.} For a continuous score, the hazard ratio represents the relative increase or decrease in the hazard of the event associated with a one-unit increase in that score. In our analysis, we compare a 0.2 unit increase in our AI test (which ranges from 0 to 1) to a 20 unit increase in the Oncotype score (which ranges from 0 to 100). A detailed explanation of how the hazard ratio is computed is in Appendix~\ref{sec:hazard-ratio}.

\paragraph{Hazard ratio of a dichotomized score.} For a dichotomized score, the patients are stratified into two categories: high-risk and low-risk. The hazard ratio is computed using a Cox proportional hazard model with the dichotomized score as the only covariate, with the low-risk group as the reference. The p-values are computed using the Wald test.

\paragraph{Multivariate Cox proportional hazards model.}

To assess the simultaneous impact of multiple variables on the outcome, we fit a multivariate Cox proportional hazards model with all variables of interest. For categorical variables in this analysis, we assign a binary indicator to all categories except one, which serves as reference.

We perform three types of multivariate analyses. In the first analysis, we include the AI test score alongside dataset indicators within the Cox model. This analysis allows us to estimate the hazard ratio associated with different features after adjusting for the varying patient outcome distributions in different datasets. We have a separate indicator variable for each dataset other than Basel, which is used as the reference group. In the second analysis, we have clinical and pathology scores within the AI test alongside the dataset indicator variables. This analysis is used to assess whether our pathology model provides prognostic information independent of the clinical model. In the final analysis, we use the AI test score along with Oncotype DX score, grade, race indicators and dataset indicators. For race, we have indicators for black, Asian and other, with white as the reference group. This analysis is performed to validate the association between our test and recurrence after adjusting for established prognostic factors. 

\paragraph{Recurrence rate as a function of the AI test score.} To assess how the AI test score corresponds to patient recurrence, we divided patients based on their AI test scores into groups using quartiles. We then used the Kaplan-Meier estimator to compute 10-year recurrence rates (with confidence intervals) across these groups of patients. The aim was to compare the average score within each group to the estimated recurrence rate within that group. This approach helped us evaluate how well the AI test score corresponded to the observed risk of recurrence, providing a clear way to assess whether higher test scores consistently correspond to higher recurrence rates.

\subsubsection{Hyperparameter selection}

In all instances, we selected the hyperparameters of the models through random search~\cite{bergstra12random}, using the training data. Unless otherwise explicitly stated we maximized the C-index.

In the hyperparameter search, the performance of pathology models was estimated using modified multiple-source cross-validation~\cite{geras13multiple}. That is, we had two collections of datasets $\mathcal{D}^T = \{D_1, D_2, \ldots, D_J\}$ and $\mathcal{D}^V = \{D_{J+1}, D_{J+2}, \ldots, D_{J+K}\}$, where each $D$ in $\mathcal{D}^T$ and in $\mathcal{D}^V$ denotes a separate cohort. Datasets in $\mathcal{D}^T$ were reserved only for training, while $\mathcal{D}^V$ could be used for training or validation. For a given hyperparameter setting, we trained $K$ models, {$m_1$, $m_2$, \ldots, $m_K$}. The training set for the $k^{\text{th}}$ model was $\mathcal{D}^T_k = \mathcal{D}^T \cup \mathcal{D}^V \backslash D_{J+k}$, and the validation set was $D_{J+k} \in \mathcal{D}^V$. The performances on the validation sets were averaged, weighted by the number of comparisons $c_k$ used in the calculation of the validation C-index using the dataset $D_{J+k}$. This procedure was used for performance estimation and hyperparameter selection as formalized in Algorithm~\ref{alg:MSCV}. To limit the variance in hyperparameter selection, for each set of hyperparameters, we trained and evaluated several models with different random seeds and averaged the validation performances.

\begin{algorithm}
    \caption{Multiple-source cross-validation for pathology models}
    \label{alg:MSCV}
    \begin{algorithmic}
        \FOR{n = $1$ to $N$}
            \STATE Sample hyperparameters $\theta_n$
            \FOR{k = $1$ to $K$}
                \STATE Train a model $m^{\theta_n}_k$ using data $\mathcal{D}^T \cup \mathcal{D}^V \backslash D_{J+k}$
                \STATE Evaluate the model $m^{\theta_n}_k$ using $D_{J+k}$ to obtain the validation performance $v^{\theta_n}_k$
            \ENDFOR
            \STATE The overall validation score of hyperparameters $\theta_n$ is $v^{\theta_n} = \frac{1}{K} \sum_{k=1}^K c_k v^{\theta_n}_k$.
            \STATE The model obtained for $\theta_n$ is $m^{\theta_n} = \frac{1}{K} \sum_{k=1}^K m^{\theta_n}_k$.
        \ENDFOR
        \STATE Select the hyperparameters $\theta^*_n$ yielding the best overall validation score.
    \end{algorithmic}
\end{algorithm}

In our implementation, when training the pathology models, we set $\mathcal{D}^T = \{\mathrm{Korea}, \mathrm{Gundersen}, \mathrm{Lithuania},\\\mathrm{BCN}, \mathrm{Malaysia}, \mathrm{METABRIC}, \mathrm{BASIS}\}$ and $\mathcal{D}^V = \{\mathrm{NYU}, \mathrm{Omica}, \mathrm{Wales}\}$.

As clinical models use much lower dimensional input than pathology models and their performance exhibits less variance, we adopted a simpler cross-validation procedure. We divided the data into $k$ splits, that is, $\mathcal{D}^C = \{D_1^C, D_2^C, \ldots, D_K^C\}$. For each iteration, we trained the models using $\mathcal{D}^C \backslash D_{k}^C$ and selected the hyperparameters that perform the best for this split based on validation performance on $D_{k}^C$. This procedure was used for performance evaluation and hyperparameter selection as formalized in Algorithm~\ref{alg:MSCV-clinical}.

\begin{algorithm}
    \caption{Multiple-source cross-validation for clinical models}
    \label{alg:MSCV-clinical}
    \begin{algorithmic}
        \FOR{k = $1$ to $K$}
            \FOR{n = $1$ to $N$}
                \STATE Sample hyperparameters $\theta_n$ and train a model $m^{\theta_n}_k$ using data $\mathcal{D}^C \backslash D_{k}^C$. 
            \ENDFOR
            \STATE Select the hyperparameters $\theta^*_k$ yielding the best overall validation score $v^{\theta_n}$ for split $D_{k}^C$.
            \STATE The model obtained for split $k$ is $m^{\theta^*_k}$
        \ENDFOR
        \STATE Obtain final model $m = \frac{1}{K} \sum_{1}^{K} m^{\theta^*_k}$
    \end{algorithmic}
\end{algorithm}

In our implementation, when training the clinical models, we trained using three splits $\mathcal{D}^C = \{D_1^C, D_2^C, D_3^C\}$. We set $D_1^C = \mathrm{NYU}, D_2^C = \mathrm{METABRIC}$, and $D_3^C = \bigcup \{ \mathrm{Korea}, \mathrm{Gundersen}, \mathrm{Lithuania}, \mathrm{BCN}, \mathrm{Malaysia}, \\\mathrm{BASIS}, \mathrm{Wales} \}$.

\subsection{Models used to build the AI test}
\label{sec:model-description}

Our AI test makes predictions using both clinical variables and digital pathology images. Internally, it is composed of many models whose predictions are ensembled in two rounds. First, the models are ensembled within their data modality. Then, the ensemble of clinical models is combined with the ensemble of digital pathology models. The intuition behind ensembling lies in the idea that different models make errors that are not perfectly correlated. Ensembling several models cancels out some of these errors, adding accuracy and robustness to predictions. An additional advantage of ensembling at the level of modalities is that for every test example, it is easy to interpret the contribution of the clinical data and the digital pathology data to the final prediction.

\subsubsection{Creating model ensembles}
\label{sec:ensembling}

We ensembled the clinical models in two steps. First, the output risk scores from all clinical models were scaled approximately to the interval \([0,1]\). That is, for each model $m_k$, we found the maximum over the entire validation set, $\mathrm{max}^k = \mathrm{max}_i\{\hat{y}^k_i\}$, and the minimum over the entire validation set, $\mathrm{min}^k = \mathrm{min}_i\{\hat{y}^k_i\}$. For the test examples, the normalized prediction from any model was computed as $y^k = \frac{\hat{y}^k - \mathrm{min}^k}{\mathrm{max}^k - \mathrm{min}^k}$. The predictions from all models were then simply averaged. That is, assuming that there were $K$ models, for any test example, the aggregated clinical prediction was computed as $y^c = \frac{1}{K} \sum_{k=1}^K y^k$.

For pathology models, as the predictions were already in the interval \([0,1]\), we simply averaged the predictions from different models. That is, for any test example, the aggregated pathology prediction was computed as $y^p = \frac{1}{K} \sum_{1}^{K} y^p_k$. We generated the final pathology model by averaging the top $K$ models across different combinations of pooling method (mean, max, or multiple instance learning), training loss function (Cox or discrete-time), and feature extractor to form the final ensemble. We used $K=10$ for both clinical and pathology models. 

Finally, the clinical score and the pathology score were averaged, that is, $y = \frac{1}{2}(y^c + y^p)$.

\subsection{Digital pathology models}

Unlike many prior approaches~\cite{amgad2024population, fernandez2022development, nimgaonkar2023development}, we did not use any hand-crafted pathomic features to build the pathology models. Instead, we obtained learned morphological features from Kestrel, our purpose-built foundation model trained on a pan-cancer dataset. These features were used to train time-to-event models. Below we present our model in greater detail.

\begin{figure}[h!]
    \centering
    \includegraphics[width=0.5\linewidth]{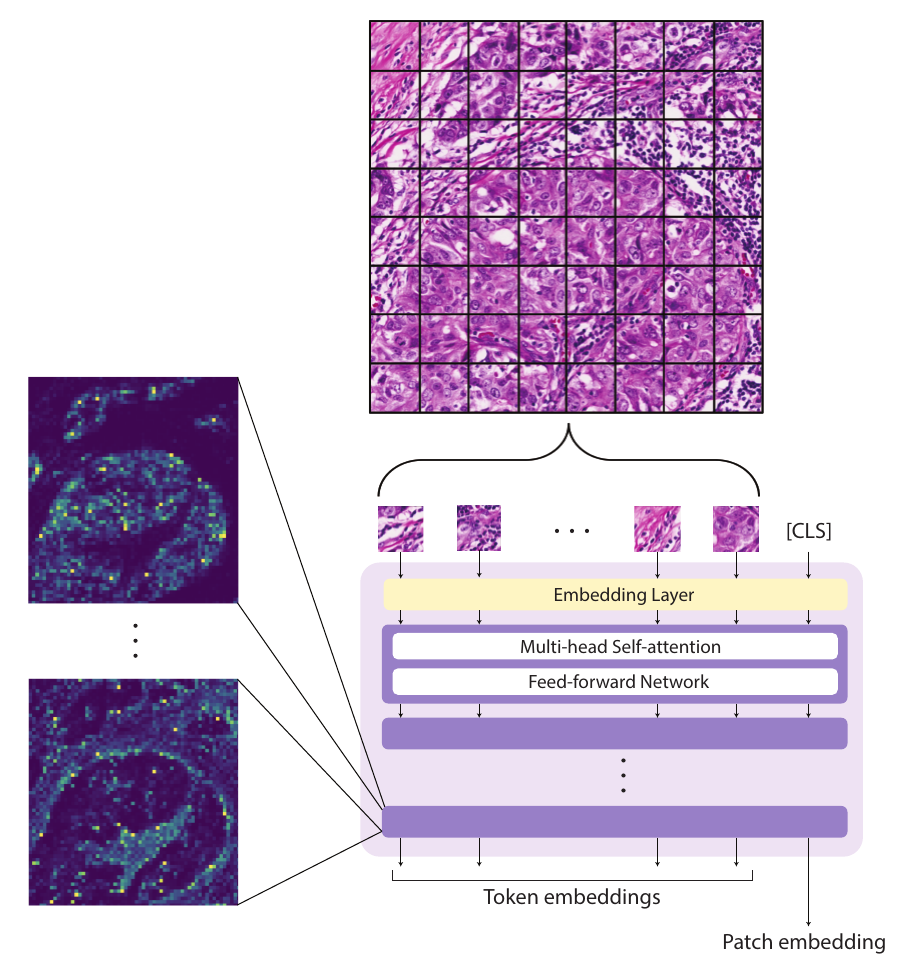}
    \caption{\textbf{Overview of the vision transformer (ViT) architecture applied to digital pathology.} Images are first broken into a grid of sub-patches, or ``tokens''. These tokens, along with a \texttt{[CLS]} token which serves to aggregate features across sub-patches, are vectorized via an embedding layer. These vectors then progress through a sequence of blocks which consist of multi-headed self-attention followed by a feed-forward network. The ViT output consists of token embeddings and an aggregate patch embedding. The patch embedding is the primary output used downstream, and both the patch and token embeddings are used for self-supervised learning. On the left, we display attention masks between the \texttt{[CLS]} embedding and token embeddings extracted from the final block. These masks provide insight into the types of features the vision transformer is using to create the patch embeddings.}
    \label{fig:vit}
\end{figure}

\begin{figure}[h!]
    \centering
    \includegraphics[width=0.8\linewidth]{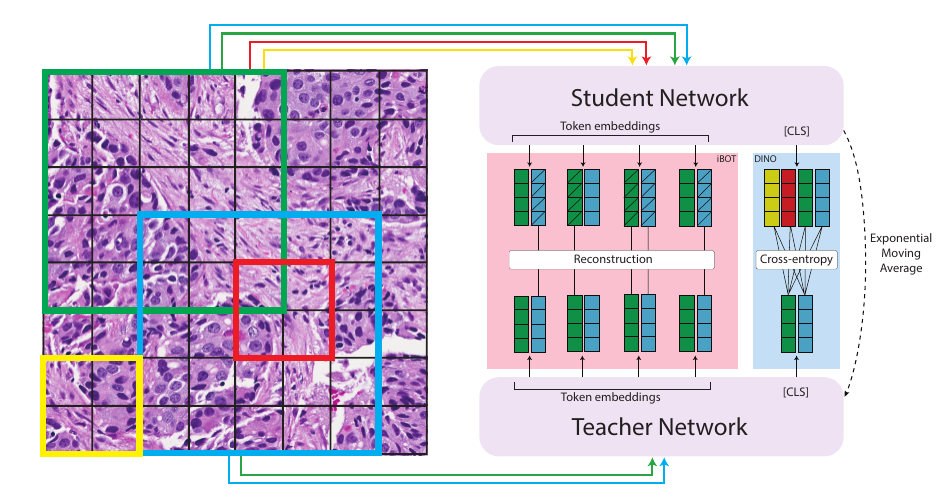}
    \caption{\textbf{Overview of self-supervised learning method DINOv2 applied to digital pathology.} First, several larger (global) and smaller (local) crops are sampled from an image and transformed with image augmentations such as blurring, rotation and scaling. Global and local crops are passed through the student network while only global crops are passed through the teacher network. For the DINO loss component, cross-entropy is computed between the student's CLS token outputs and teacher's CLS token outputs. For the iBOT loss component, input tokens are randomly masked before passing global crops through the student network, and a reconstruction loss is computed between the corresponding student network token embeddings and the corresponding teacher network token embeddings.}
    \label{fig:dinov2}
\end{figure}

\paragraph{Kestrel's architecture.}
Kestrel is a ViT-L which contains 303M parameters. An overview of the vision transformer architecture is provided in Figure~\ref{fig:vit}.

\paragraph{Preprocessing histopathology slides.} Digital pathology slides are extremely high resolution, often exceeding 1 billion pixels per image. Additionally, large regions of these images are empty. Therefore, to extract only meaningful tissue, we used Otsu's method~\cite{otsu1979threshold} to distinguish background from foreground. Subsequently, we patchify the foreground into non-overlapping $256\times256$ patches at 20$\times$ magnification (0.5 microns per pixel). Thus, each slide is reduced to a set of patches. There are typically between 8,000 to 16,000 patches per slide.

\paragraph{Training Kestrel.} Kestrel~\cite{cappadona2024squeezing} was trained using a self-supervised learning method DINOv2~\cite{oquab2024dino} on 400M patches extracted from 45,000 histopathology slides. We tuned the hyperparameters for training Kestrel by evaluating trained models on a suite of 16 benchmarks. These benchmarks come mostly from public challenges and are designed to test a model's ability to learn to recognize different characteristics of the micro-tumor environment, for example, tumor cellularity and histology. This has enabled us to train models that excel at generating meaningful features from tissue patches.

\paragraph{Extracting digital pathology features.} Slides were preprocessed to extract only foreground patches. These foreground patches were passed through our foundation model to obtain patch-level embeddings. These patch-level embeddings are used as inputs to a time-to-event model.

Our model's predictions were generated using H\&E-stained slides from the same tissue block that was previously used for Oncotype~DX testing. For an overwhelming majority of cases (approximately 93\% of the patients across all evaluation cohorts), we used only a single slide. For the remaining minority of the cases for which we had more than one slide, we averaged the predictions across slides.

\paragraph{Training the digital pathology models}

We trained two types of models using features extracted by Kestrel as input: Cox proportional hazard models and discrete-time models.

\subsubsection{Cox proportional hazard model}
Consider a model, $m$, parameterized by $\beta$, with the training dataset $D_T$ and the validation dataset $D_V$. The training loss for this model is computed as 

\begin{equation*}
\mathcal{L} = (1-\alpha)\mathcal{L}_{\text{Cox}} + \alpha\mathcal{R}(\beta),
\end{equation*}
where $\mathcal{L}_{\text{Cox}}$ is the negative log of the partial likelihood computed for the training data, $\mathcal{R}$ is a regularizer, and $\alpha$ is a hyperparameter controlling the the relative weight of $\mathcal{L}_{\text{Cox}}$ and $\mathcal{R}$.

Specifically, the training loss $\mathcal{L}_{\text{Cox}}$ is computed as 
\begin{equation*}
    \mathcal{L}_{\text{Cox}} = \frac{1}{|D_T|} \sum_{i \in D_T} \delta_i \log\left(\sum_{j \in \mathcal{S}_i} \exp[g(x_j)-g(x_i)]\right),
\end{equation*}
where $x_i$ represents image features for a whole slide image, $g$ is a neural network transforming $x_i$, $\delta_i$ indicates whether the patient experienced an event, and $\mathcal{S}_i$ represents the risk set (the individuals that have not yet experienced an event at time $t_i$).

The regularizer $\mathcal{R}$ is computed as
\begin{equation*}
    \mathcal{R}(\beta) =  \frac{(1-\gamma)}{2} ||\beta||_2^2 + \gamma ||\beta||_1,
\end{equation*}
where $\gamma$ is a hyperparameter controlling the relative weight of the L1 and L2 regularization terms. The entire training loss is optimized with Adam~\cite{kingma2015adam}, a variant of stochastic gradient descent.

\subsubsection{Discrete-time model}

We divide times from $t_0$ to $t_J$ into $J$ contiguous time intervals $(t_0,t_1], (t_1,t_2], ..., (t_{J-1},t_J]$, where $t_0=0$ and $t_J = \infty$.

For subject $i$ with covariates $x_i$ (here, features extracted from a whole slide image), the hazard in interval $A_j = (t_{j-1}, t_j]$ is computed as $$\lambda_{ij}(x_i) = \Pr(T_i \in A_j | T_i > t_{j-1}, x_i),$$ and survival probability is computed as
$$S_i(t | x_i) = \Pr (T_i > t | x_i) = \prod_{j: t_j \leq t} (1-\lambda_{ij}(x_i)).$$

We introduce an indicator $d_{ij} = \mathbbm{1}[T_i \in A_{ij}] = \mathbbm{1}[t_{j-1} < T_i \leq t_j]$, which for censored subjects is given by $(d_{i1}, \ldots, d_{ij_i}) = (0, \ldots, 0)$ and for subjects that experience the event is given by $(d_{i1}, \ldots, d_{ij_i}) = (0, \ldots, 0, 1)$. The likelihood can then be defined as 
\begin{align*}
\mathcal{L_{\text{DT}}} = \prod_{i=1}^n \lambda_{i{j_i}}(x_i)^{d_{i{j_i}}} \prod_{j=1}^{j_i-1} (1 - \lambda_{ij}(x_i))^{1-d_{ij}},
\end{align*}
and the log-likelihood is
$$\log \mathcal{L_{\text{DT}}} = \sum_{i=1}^n  \left[  d_{i{j_i}} \log \lambda_{i{j_i}}(x_i) +   \sum_{j=1}^{j_i} (1-d_{ij}) \log(1 - \lambda_{ij}(x_i)) \right].
$$

Then, we can model the score in the $j$\textsuperscript{th} time interval using a neural network as $\phi_j(x_i)$. To put the hazard in the $[0, 1]$ interval, we set 
\begin{align*}
\lambda_{ij}(x_i) = \frac{1}{1+\exp [- \phi_j(x_i)]}.
\end{align*}
We can estimate $\phi_j(x_i)$ by minimizing the negative log-likelihood. Regularization is added to $\mathcal{L_{\text{DT}}}$ the same way it is added to $\mathcal{L_{\text{Cox}}}$ described in the previous section.

\subsubsection{Aggregation of patch embeddings}

In order to train time-to-event neural networks with Cox or discrete-time losses, it is necessary to aggregate patch embeddings into a single vector representation that can be passed into a standard feed-forward neural network. We used two strategies to do this: parameterless pooling and attention-weighted multiple instance learning (MIL) pooling.

\paragraph{Parameterless pooling.} It has been shown that simple forms of pooling such as mean- and max-pooling perform well on digital pathology tasks where aggregation is needed. Thus, one set of the trained time-to-event models in our pathology ensemble consists of networks trained on mean- or max-pooled patch embeddings with a Cox loss.

\paragraph{Attention-weighted MIL.} It has been shown that simple attention-based networks, such as gated attention \cite{ilse2018attention}, are also effective ways of aggregating patch embeddings in digital pathology. Thus, we trained a complementary set of time-to-event models using a gated attention network with a discrete-time loss.

\subsection{Clinical models}
\label{sec:clinical-model-description}
For building a model for clinical variables, we utilized CatBoost~\cite{prokhorenkova2018catboost}, a gradient boosting decision tree ensemble. Its primary advantage over other possible models for this application, e.g., neural networks, is how it handles categorical variables which are inevitable in clinical data. Here, CatBoost is specifically trained with the AFT (accelerated failure time) loss \cite{wei1992accelerated}. We utilized AFT models with three possible distributions (selected in hyperparameter search): normal, logistic, and extreme value.

We used eight routinely collected variables (see Table \ref{tab:clinical_values}) that characterize breast cancer and are typically used to guide treatment decisions. For instance, ER and PR statuses help determine tumor hormone receptor sensitivity, which is critical for hormone-based therapies. HER2 status is important for targeted therapies such as trastuzumab, which is effective in HER2+ cancers.

\begin{table}[ht]
\caption{Possible values for the clinical variables used as inputs for our clinical model.}
\label{tab:clinical_values}
\centering
\begin{tabular}{|l|l|}
\hline
\textbf{clinical variable}            & \textbf{possible values}                 \\ \hline
patient age                           & [0, 100] \\\hline
estrogen receptor (ER) status         & positive, negative, unknown                       \\
progesterone receptor (PR) status     & positive, negative, unknown                       \\ 
HER2 biomarker status                 & positive, negative, equivocal, unknown            \\ 
\hline
T stage (tumor size/extent)           & T1mi, T1a, T1b, T1c, T2, T3, T4, TX, unknown           \\
N stage (nodal involvement)           & N0, N1, N2, N3, NX, unknown                         \\ 
\hline
has IDC component               & yes, no, unknown \\

has ILC component                                    & yes, no, unknown        \\ \hline
\end{tabular}
\end{table}

\bibliography{references}
\bibliographystyle{ieeetr}

\section*{Data availability}

TCGA-BRCA data can be accessed from The Cancer Imaging Archive under doi:10.7937/K9/TCIA.2016.AB2NAZRP. The BASIS dataset can be requested from the Wellcome Sanger Institute and was accessed from the European Genome-Phenome Archive (EGA) under accession number EGAS00001001178. The METABRIC dataset can be requested from the METABRIC Committee and was accessed from EGA under accession number EGAD00010000270. The Providence dataset can be accessed from Nightingale via doi:/10.48815/N5159B. The remaining datasets are private or proprietary and are not publicly available.

\section*{Code availability}

The model can be accessed for non-commercial research. We have also documented all experiments with sufficient details in our Methods section to enable independent replication. Although the codebase cannot be shared due to its proprietary nature, the core components of our work rely on open-source repositories:

\begin{itemize}
    \item DINOv2 model architecture for self-supervised training:
          \begin{itemize}
              \item \url{https://github.com/facebookresearch/dinov2}
          \end{itemize}
    
    \item PyTorch library for training and inference:
          \begin{itemize}
              \item \url{https://github.com/pytorch/pytorch}
          \end{itemize}
    
    \item For preprocessing whole slide images:
          \begin{itemize}
              \item OpenSlide: \url{https://openslide.org}
              \item wsidicom: \url{https://github.com/imi-bigpicture/wsidicom}
          \end{itemize}
    
    \item For training and models for survival analysis:
          \begin{itemize}
              \item CatBoost: \url{https://github.com/catboost/catboost}
              \item PyCox: \url{https://github.com/havakv/pycox}
              \item Scikit-learn: \url{https://github.com/scikit-learn/scikit-learn}
              \item Torch Tuples: \url{https://github.com/havakv/torchtuples}
          \end{itemize}
    
    \item For result analysis:
          \begin{itemize}
              \item NumPy: \url{https://github.com/numpy/numpy}
              \item Metagen: \url{https://cran.r-project.org/web/packages/meta/meta.pdf}
              \item Lifelines: \url{https://github.com/CamDavidsonPilon/lifelines}
              \item Concordance: \url{https://cran.r-project.org/web/packages/survival/vignettes/concordance.pdf}
          \end{itemize}
    
    \item For visualizations:
          \begin{itemize}
              \item Seaborn: \url{https://github.com/mwaskom/seaborn}
              \item Matplotlib: \url{https://github.com/matplotlib/matplotlib}
          \end{itemize}
\end{itemize}

\section*{Acknowledgements}

DZP acknowledges funding from the Research Council of Lithuania (LMTLT), agreement No. S-PD-22-86.
For METABRIC cohort: This study makes use of data generated by the Molecular Taxonomy of Breast Cancer International Consortium. Funding for the project was provided by Cancer Research UK and the British Columbia Cancer Agency Branch.
For TCGA-BRCA cohort: The results shown here are in whole or part based upon data generated by the TCGA Research Network: http://cancergenome.nih.gov/.
For Wales cohort: Biosamples were obtained from the Wales Cancer Bank (doi:10.5334/ojb.46) which is funded by Health and Care Research Wales. Other investigators may have received specimens from the same subjects.
For Breast Cancer Now Biobank cohort: the authors wish to acknowledge the roles of the Barts Cancer Now Tissue Bank in collecting and making available the samples and/or data, and the patients who have generously donated their tissues and shared their data to be used in the generation of this publication.

\section*{Author contributions}

JW and KJG contributed to study conception. CFG, LP, YLC and KJG contributed as research advisors. NC, FS, UO, VS, LMP, NK, AB and FJE provided clinical guidance. JW, KZ, JC and KL wrote code, developed infrastructure and trained models throughout the study. JW, KZ, JC, JE, EDC, YJK, FH, ZS, NT, MS, FY, EL, TH, BS, DR, AC, VS, MV, LS, SS, DB, KC, YZ, LD, AS, WA, RB, JWP, HM, SHT, VA, DZP, AL, BDP, CB, SYJ, JPY, SHL and KJG worked on data preparation. JW, KZ, JC, IO, CFG and KJG performed evaluation and analysis. IO contributed as a statistical advisor. JW, KZ, JC, IO, CFG and KJG worked on drafting and revising the manuscript. All authors critically reviewed the paper and the results and approved the final version.

\section*{Competing interests}

JW, KZ, JC, CFG, FS, ZS, AB, FJE, YLC and KJG are equity holders of Ataraxis AI. IO served as a consultant for Ataraxis AI. New York University (NYU) maintains financial and intellectual property interests in Ataraxis AI that are pertinent to the research presented in this manuscript. JW and KJG are inventors on a US patent application filed corresponding to some of the methodological aspects of this work. The remaining authors declare no competing interests.

\newpage
\appendix

\section{Appendix}
\counterwithin{figure}{section}
\counterwithin{table}{section}

\subsection{Concordance index}
\label{sec:c-index}

C-index, originally proposed by Harrell et al.~\cite{harrell1982evaluating}, is a metric of accuracy for time-to-event models, analogous to AUROC for classification, adjusted for the possibility of censoring. It is defined as the ratio of the number of \textit{concordant} pairs of test examples to the number of all \textit{comparable} pairs of test examples. A test example is defined by its event time, $T$, its vector of input features, $x$, and $\delta$ that indicates the presence of an event. Two test examples are comparable if the example with the shorter event time is not censored (i.e., experienced an event). Two test examples are concordant if the model assigns a higher risk score to the example with a shorter event time. Given $\delta_i = 1$ indicating the presence of an event and $T_i$ the `time-to-event' response for example $i$, we use Equation~\ref{eq:c-index} to compute the C-index:

\begin{equation}
    \mathrm{C} = \frac{\sum_{i, j} \Big[\text{risk}(x_i)>\text{risk}(x_j)\Big] \cdot \Big[T_i < T_j\Big] \cdot \delta_i}{\sum_{i, j} \Big[T_i < T_j\Big] \cdot \delta_i}.
    \label{eq:c-index}
\end{equation}

\subsection{Hazard ratio}
\label{sec:hazard-ratio}

Hazard ratio is a metric appropriate for measuring how well a model can divide patients into two groups, e.g., low- and high-risk. The hazard ratio is computed using the Cox proportional hazards model, in which the input indicates one of the two groups. That is, we model the hazard for an individual at time $t$ as $\lambda(t|x) = \lambda_0(t)\exp{(\beta x)}$, where $x$ is the group indicator ($x = 0$ or $x = 1$), $\lambda_0(t)$ is the baseline hazard function which represents the hazard when $x = 0$ and $\beta$ is the regression coefficient associated with the variable that indicates the group a patient belongs to. Then, based on the data, this model is fitted using partial likelihood. For the fitted model, hazard ratio is computed as 

$$
\mathrm{HR} = \frac{\lambda(t | x = 1)}{\lambda(t | x = 0)}
            = \frac{\lambda_0(t)\exp{(\beta \cdot 1)}}{\lambda_0(t)\exp{(\beta \cdot 0)}}
            = \exp{(\beta)}.
$$

\noindent
If $\mathrm{HR} > 1$, the group with $x = 1$ has a higher hazard of experiencing the event in comparison to the group with $x = 0$. If $\mathrm{HR} < 1$, the group with $x = 0$ has a higher hazard of experiencing the event compared to the group with $x = 1$. For example, if $\mathrm{HR} = 0.5$ the risk for the group with $x = 0$ is twice the risk for the group with $x = 1$.

Hazard ratio can be defined analogously for a continuous $x$. Then, it is interpreted as the change in the hazard associated with a one-unit increase in $x$. That is,

$$
\mathrm{HR} = \frac{\lambda(t | x = \alpha + 1)}{\lambda(t | x = \alpha)}
            = \frac{\lambda_0(t)\exp{(\beta \cdot (\alpha + 1))}}{\lambda_0(t)\exp{(\beta \cdot \alpha)}}
            = \exp{(\beta)}.
$$

Additionally, the logrank test~\cite{mantel1966evaluation} is commonly utilized to determine if the hazard ratio is significantly different from one between two groups. The Wald test is used to assess the significance of individual predictors or the hazard ratio for a specific variable within a multivariate Cox model.

\subsection{Random effects model}
\label{sec:random_effects}

When comparing metrics across different datasets, a random effects model can be applied to account for the variability between these datasets. The random effects model accounts for both within-dataset variability (the inherent variation in the evaluated metric within each dataset) and between-dataset variability (differences between the datasets themselves).

The random effects model takes the following form
$$M_i = \mu + u_i + \epsilon_i,$$
where $M_i$ is the metric for dataset $i$, $\mu$ is the overall effect, $u_i$ is the random effect for dataset $i$ ($u_i \sim \mathcal{N}(0, \tau^2)$, $\tau^2$ represents the variance between datasets) and $\epsilon_i$ is the within-data-set noise ($\epsilon_i \sim \mathcal{N}(0, 
\sigma^2)$, $\sigma^2$ represents within-dataset variance). The overall fixed effect $\mu$ provides a summary of the model's average performance across datasets. The model is fit using maximum likelihood, as implemented in the Metagen package~\cite{schwarzer2007meta}.

\subsection{Endpoint definitions}
\label{appendix:endpoints}

We selected disease-free interval (DFI) (as defined in Table~\ref{tab:endpoint_definitions}) as the primary endpoint because it encompasses a broad range of breast cancer-related outcomes. We deemed overall survival and disease-free survival, both of which includes death as part of the endpoint, to be less relevant to our work since not all deaths are related to breast cancer. Ideally, we could exclude all deaths unrelated to breast cancer, but retrospectively determining the cause of death is challenging. Similarly, we deemed distant recurrence to be less informative as it excludes local-regional recurrences, which we also consider to be clinically significant outcomes. 

\begin{table}[ht]
    \centering
    \caption{Events (local recurrence, distant recurrence, and/or death from any causes) corresponding to each endpoint.}
    \resizebox{\textwidth}{!}{
    \begin{tabular}{l|p{3.15cm}|p{3.15cm}|p{3.15cm}}
         \textbf{endpoint} & \textbf{local-regional recurrence} & \textbf{distant recurrence} & \textbf{death (any cause)} \\
         \hline
         overall survival (OS) & & & \checkmark \\
         disease-free interval (DFI) & \checkmark & \checkmark & \\
         distant recurrence-free interval (DRFI) & & \checkmark &  \\
         recurrence-free survival (RFS) & \checkmark & \checkmark & \checkmark \\
         distant recurrence-free survival (DRFS) & & \checkmark & \checkmark \\
    \end{tabular}}
    \label{tab:endpoint_definitions}
\end{table}

\subsection{Multivariate Cox analyses}
\label{appendix:cox-multivariate-analysis}

Multivariate analysis was conducted to assess the significance of the pathology model adjusted for the clinical model (see Table \ref{tab:cox-path-clin}), importance of AI test after adjusting for datasets (see Table \ref{tab:cox-multimodal}), and AI test after adjusting for various additional factors (see Table \ref{tab:cox-oncotype-multimodal}). 

\begin{table}[htbp]
    \centering
    \caption{Cox multivariate analysis was performed across all test sets, evaluating the hazard ratio associated with the clinical and pathology model scores with DFI as the endpoint. Separate analyses were conducted for the entire patient cohort and for the subset of patients with Oncotype scores. Dataset indicator variables were included to adjust for differences between datasets.}
    \label{tab:cox-path-clin}
\input{figures/plots_for_paper_updated20241027/cox_multivariate_pooled_datasets/clinical_ensemble_x5_pathology_ensemble_x5_dataset_indicator}
\end{table}
\begin{table}[htbp]
    \centering
    \caption{Cox multivariate analysis was conducted across all test sets, evaluating the hazard ratios associated with the AI test with Disease-Free Interval (DFI) as the primary endpoint. Dataset-specific indicators were included to adjust for inter-dataset variability in outcomes.}
    \label{tab:cox-multimodal}
    \input{figures/plots_for_paper_updated20241027/cox_multivariate_pooled_datasets/combined_models}
\end{table}

\begin{table}[]
    \centering
    \caption{Cox multivariate analysis was conducted across all test sets with Oncotype scores, evaluating the hazard ratios associated with the AI test, cancer grade, age, and race, with Disease-Free Interval (DFI) as the primary endpoint. Separate analyses were performed with and without the inclusion of the AI score. Dataset-specific indicators were included to adjust for inter-dataset variability in outcomes. To ensure a fair comparison of hazard ratios, we scale both AI test score and Oncotype to a common range of [0-5]. Oncotype, originally on a 0-100 scale, is divided by 20, while our score, on a 0-1 scale, is multiplied by 5.}
    \label{tab:cox-oncotype-multimodal}
\input{figures/plots_for_paper_updated20241027/cox_multivariate_pooled_datasets/grade_race_age_oncotype_ai_test}
\end{table}

\subsection{Forest plots for all endpoints}
\label{appendix:forest-plots-all-endpoints}

\begin{figure}[htbp]
    \centering
    \begin{subfigure}{0.45\textwidth}
        \caption{DFI (disease free interval)}
        \centering
        \includegraphics[trim={180 160 270 210}, clip, width=\linewidth]{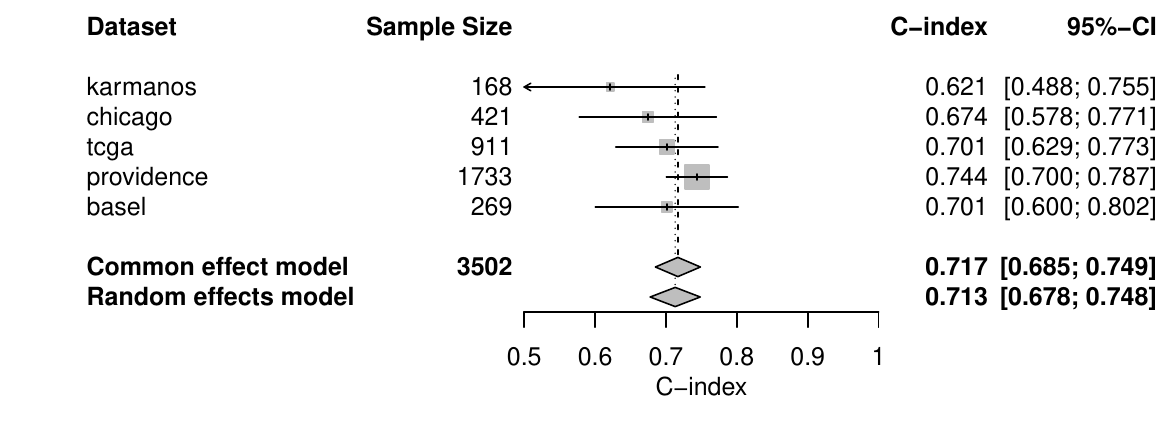}
    \end{subfigure}
    \begin{subfigure}{0.45\textwidth}
        \caption{DRFI (distant disease free interval)}
        \centering
        \includegraphics[trim={180 160 270 210}, clip, width=\linewidth]{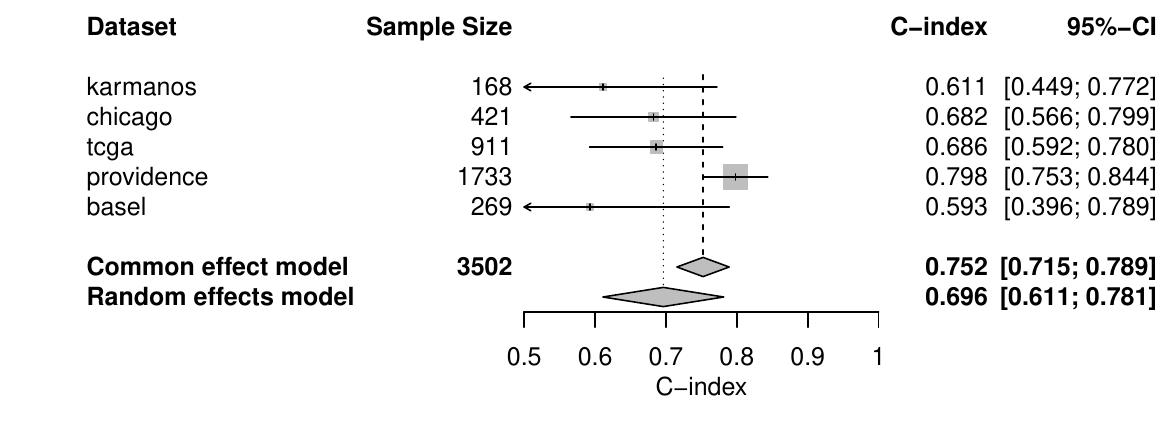}
    \end{subfigure}
    \begin{subfigure}{0.45\textwidth}
        \caption{RFS (recurrence free survival)}
        \centering
        \includegraphics[trim={180 160 270 210}, clip, width=\linewidth]{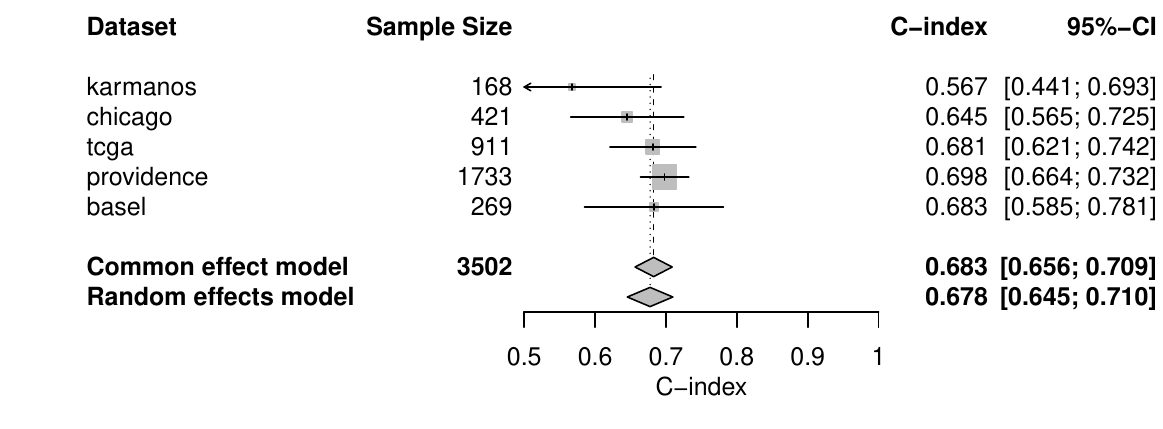}
    \end{subfigure}
    \begin{subfigure}{0.45\textwidth}
        \caption{DRFS (distant recurrence free survival)}
        \centering
        \includegraphics[trim={180 160 270 210}, clip, width=\linewidth]{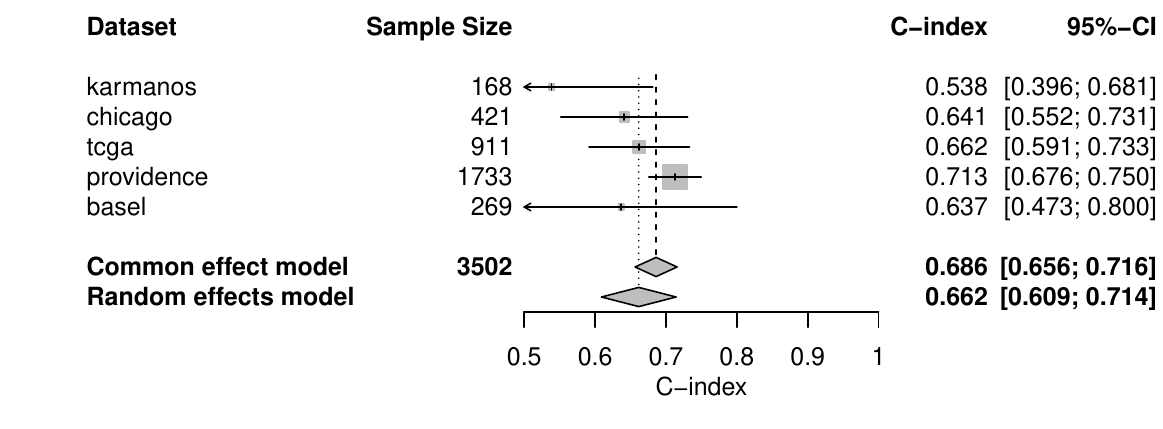}
    \end{subfigure}
    \begin{subfigure}{0.45\textwidth}
        \caption{OS (overall survival)}
        \centering
        \includegraphics[trim={180 160 270 210}, clip, width=\linewidth]{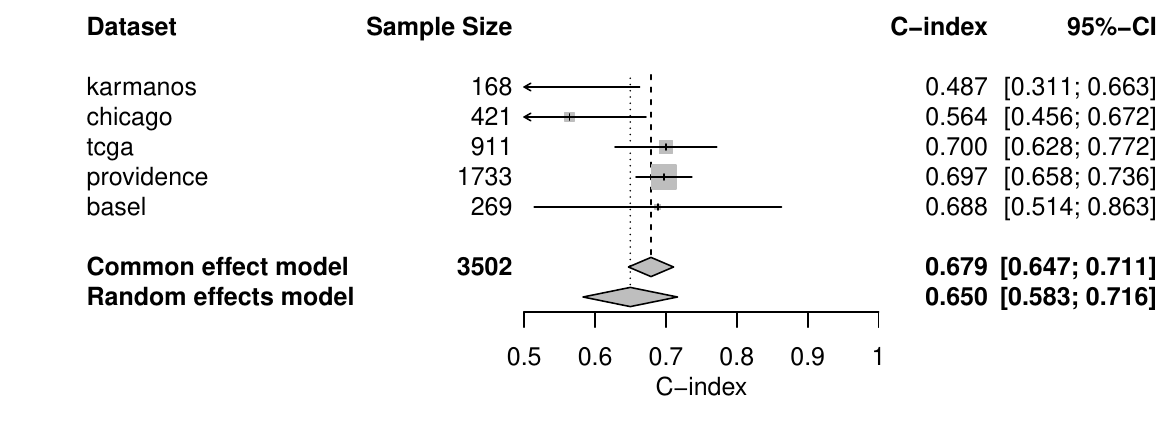}
    \end{subfigure}
    \caption{Random effect models for aggregating C-index based on continuous AI score for DFI and secondary endpoints.}
    \label{fig:mixed-effect-subgroups-all-endpoints-cindex}
\end{figure}

\begin{figure}[htbp]
    \centering
    \begin{subfigure}{0.45\textwidth}
        \caption{DFI (disease free interval)}
        \centering
        \includegraphics[trim={180 160 260 210}, clip, width=\linewidth]{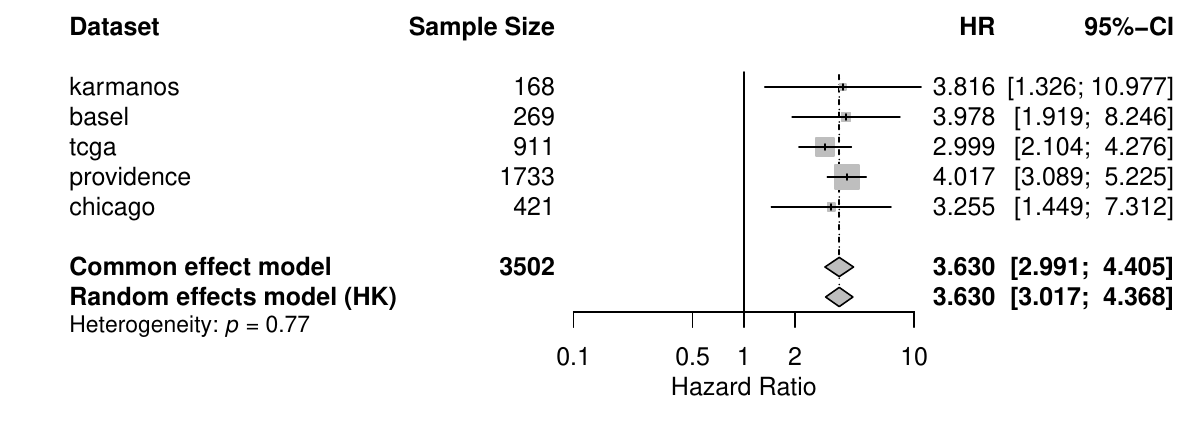}
    \end{subfigure}
    \begin{subfigure}{0.45\textwidth}
        \caption{DRFI (distant disease free interval)}
        \centering
        \includegraphics[trim={180 160 260 210}, clip, width=\linewidth]{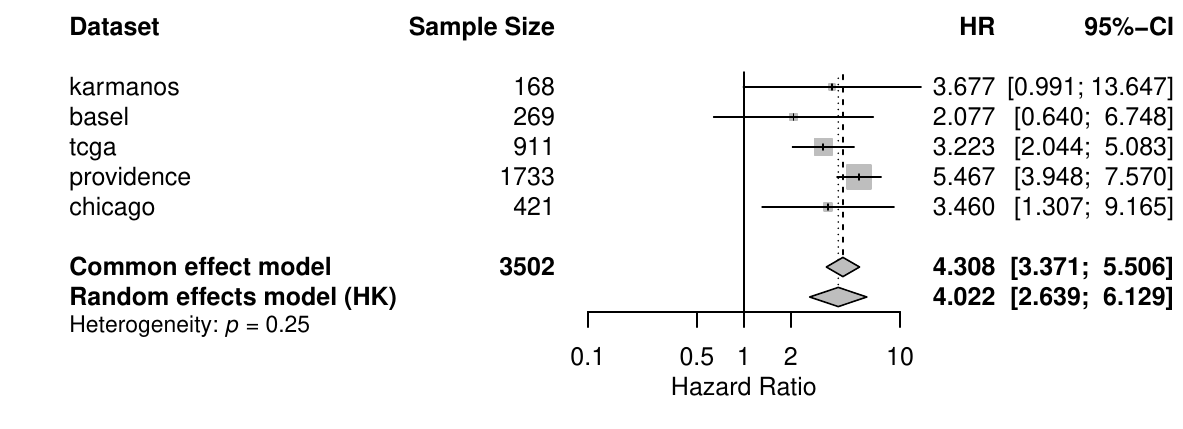}
    \end{subfigure}
    \begin{subfigure}{0.45\textwidth}
        \caption{RFS (recurrence free survival)}
        \centering
        \includegraphics[trim={180 160 260 210}, clip, width=\linewidth]{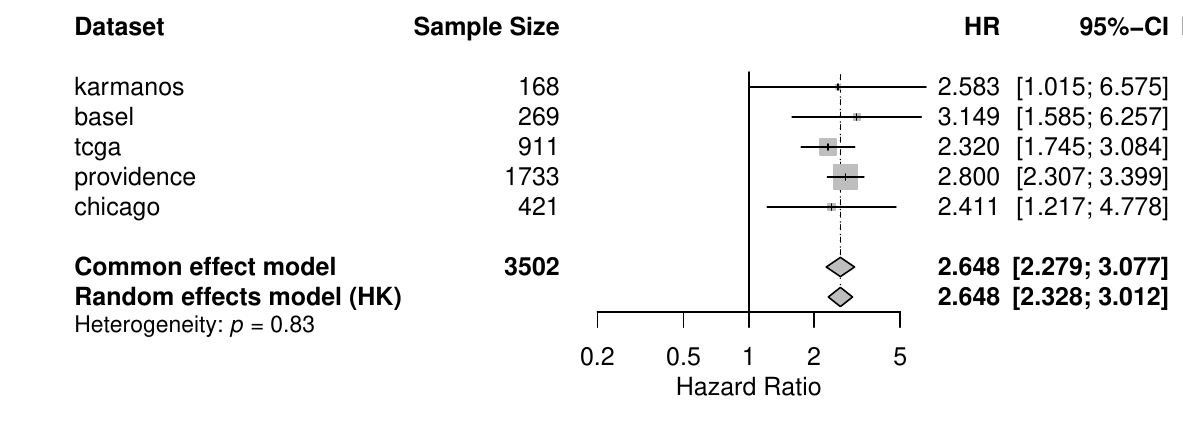}
    \end{subfigure}
    \begin{subfigure}{0.45\textwidth}
        \caption{DRFS (distant recurrence free survival)}
        \centering
        \includegraphics[trim={180 160 260 210}, clip, width=\linewidth]{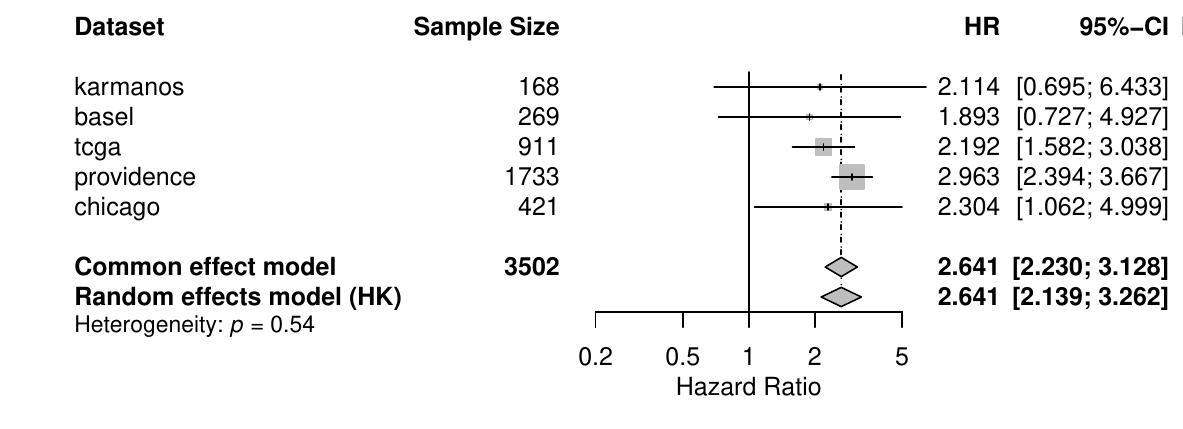}
    \end{subfigure}
    \begin{subfigure}{0.45\textwidth}
        \caption{OS (overall survival)}
        \centering
        \includegraphics[trim={180 160 260 210}, clip, width=\linewidth]{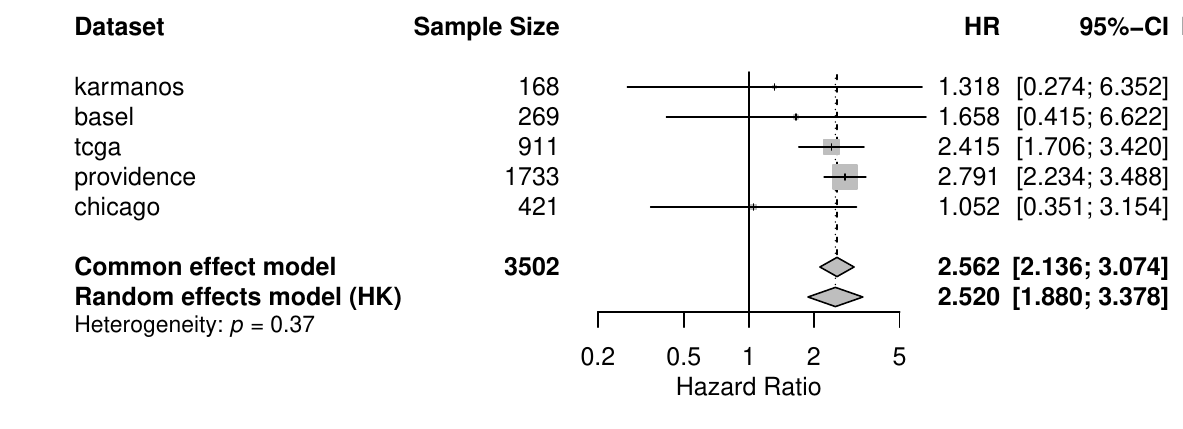}
    \end{subfigure}
    \caption{Random effect models for aggregating hazard ratios based on continuous AI score for DFI and secondary endpoints.}
    \label{fig:mixed-effect-subgroups-all-endpoints-hr}
\end{figure}

Here, we present complete results for primary and exploratory endpoints. C-index is shown in Figure \ref{fig:mixed-effect-subgroups-all-endpoints-cindex} and hazard ratios is shown in Figure \ref{fig:mixed-effect-subgroups-all-endpoints-hr}. These results include our model’s performance across various endpoints, the performance of the model on individual datasets along with pooled outcomes derived from a random effects model. 

\subsection{Forest plots for clinically relevant subgroups}
\label{appendix:forest-plots-all-subgroups}

\begin{figure}[htbp]
    \centering
    \begin{subfigure}{0.45\textwidth}
        \caption{All patients}
        \centering
        \includegraphics[trim={180 170 270 210}, clip, width=\linewidth]{figures/plots_for_paper_updated20241027/cindex_subgroup_results/DFI/global/Multimodal_C-index.pdf}
    \end{subfigure}
    \begin{subfigure}{0.45\textwidth}
        \caption{HER2+}
        \centering
        \includegraphics[trim={180 180 270 220}, clip, width=\linewidth]{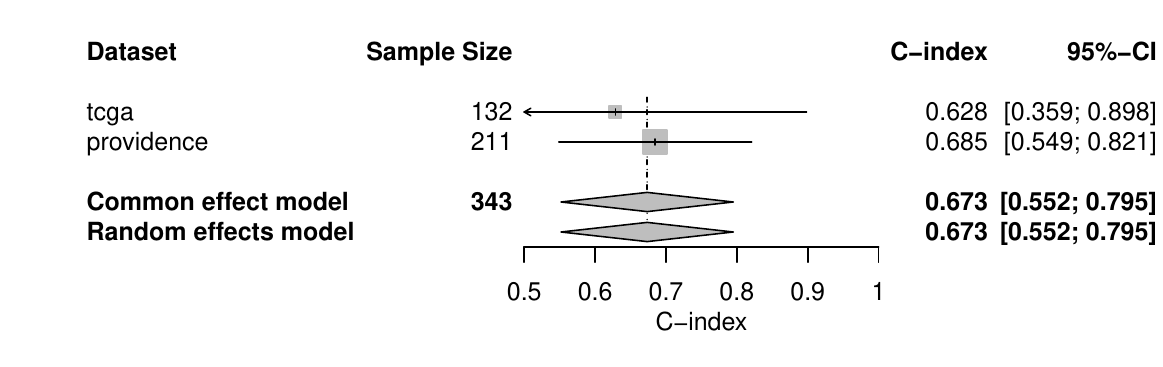}
    \end{subfigure}
    \begin{subfigure}{0.45\textwidth}
        \caption{TNBC}
        \centering
        \includegraphics[trim={180 180 270 220}, clip, width=\linewidth]{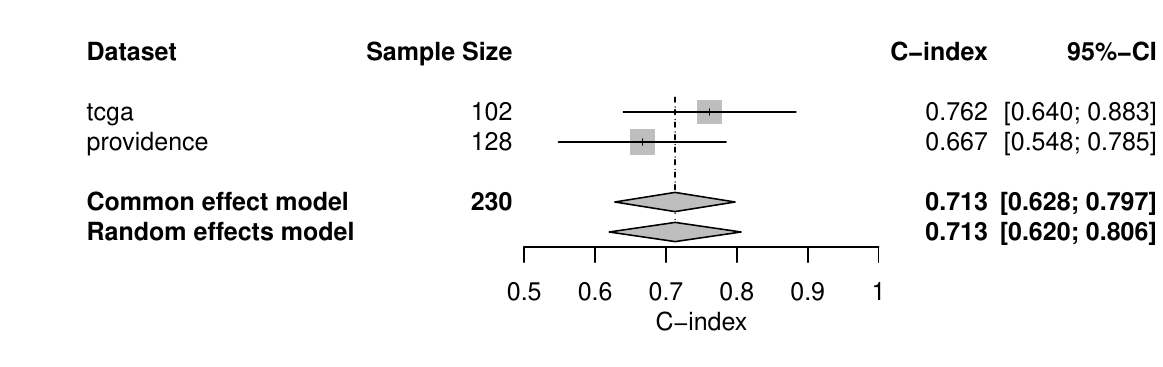}
    \end{subfigure}
    \begin{subfigure}{0.45\textwidth}
        \caption{HR+ HER2\textminus\ }
        \centering
        \includegraphics[trim={180 170 270 200}, clip, width=\linewidth]{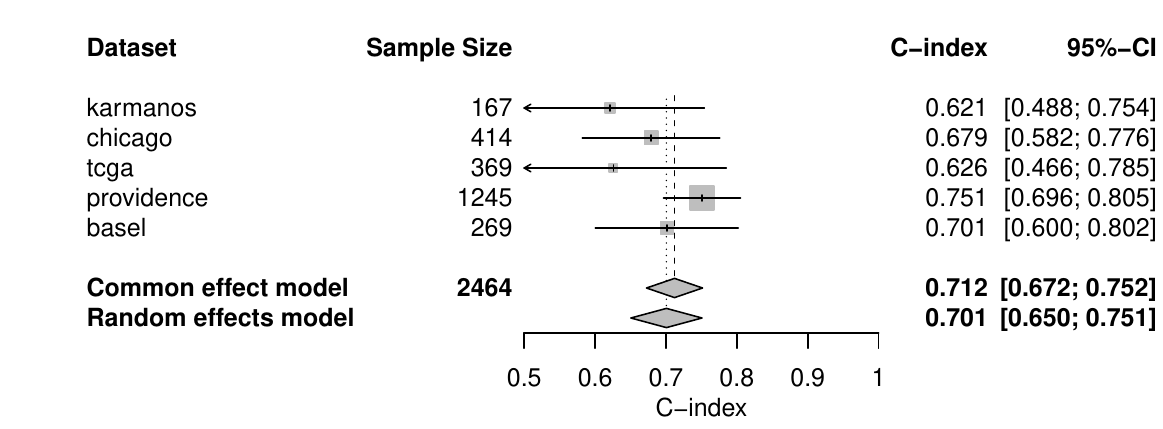}
    \end{subfigure}
    \begin{subfigure}{0.45\textwidth}
        \caption{HR+ HER2\textminus\ ET only }
        \centering
        \includegraphics[trim={180 180 270 220}, clip, width=\linewidth]{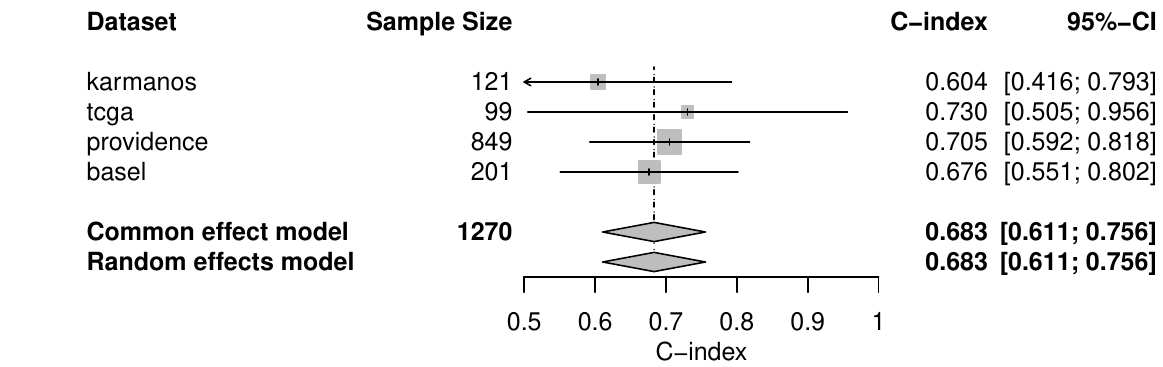}
    \end{subfigure}
    \begin{subfigure}{0.45\textwidth}
        \caption{HR+ HER2\textminus\ ET+CT}
        \centering
        \includegraphics[trim={180 180 270 200}, clip, width=\linewidth]{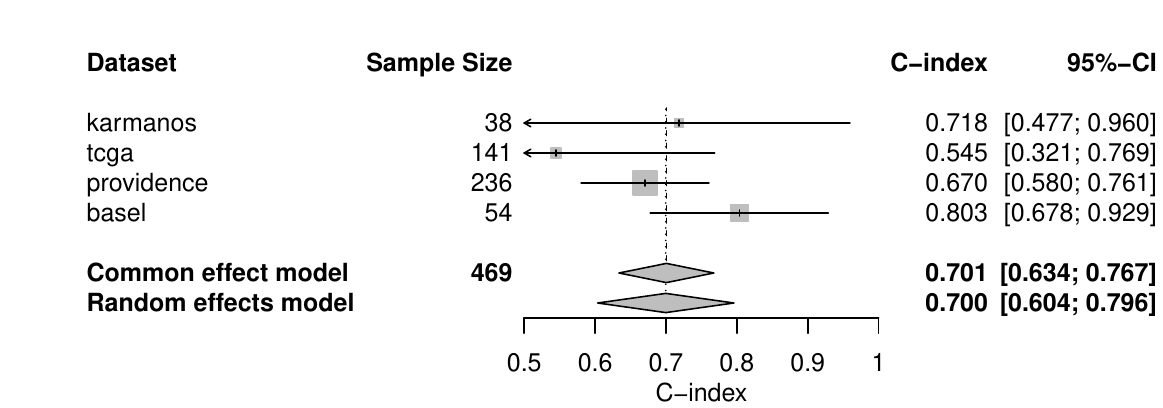}
    \end{subfigure}
    \caption{Random effect models for aggregating C-index based on continuous AI score for different patient subgroups.}
    \label{fig:mixed-effect-subgroups-dfi-cindex}
\end{figure}

\begin{figure}[htbp]
    \centering
    \begin{subfigure}{0.45\textwidth}
        \caption{All patients}
        \centering
        \includegraphics[trim={180 170 200 200}, clip, width=\linewidth]{figures/plots_for_paper_updated20241027/forest_plots_hr/hazard_ratios_continuous_multimodal_score/DFI/global/hazard_ratio_forest_plot.pdf}
    \end{subfigure}
    \begin{subfigure}{0.45\textwidth}
        \caption{HER2+}
        \centering
        \includegraphics[trim={180 180 200 200}, clip, width=\linewidth]{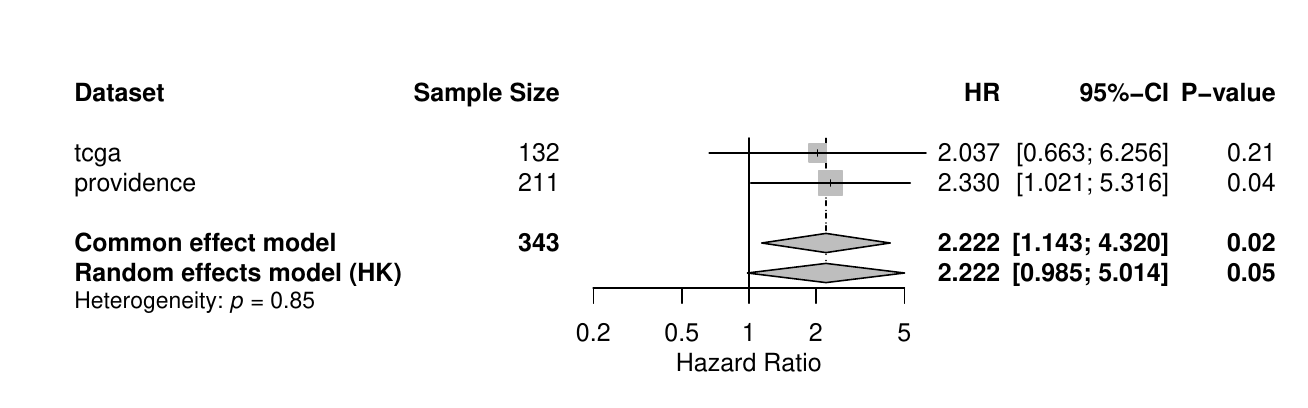}
    \end{subfigure}
    \begin{subfigure}{0.45\textwidth}
        \caption{TNBC}
        \centering
        \includegraphics[trim={180 180 200 200}, clip, width=\linewidth]{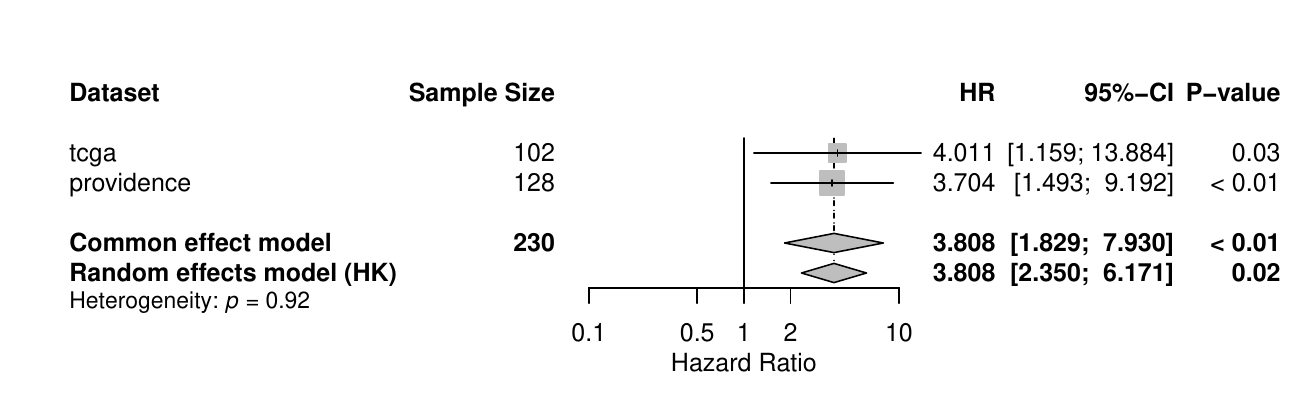}
    \end{subfigure}
    \begin{subfigure}{0.45\textwidth}
        \caption{HR+ HER2\textminus\ }
        \centering
        \includegraphics[trim={180 170 200 200}, clip, width=\linewidth]{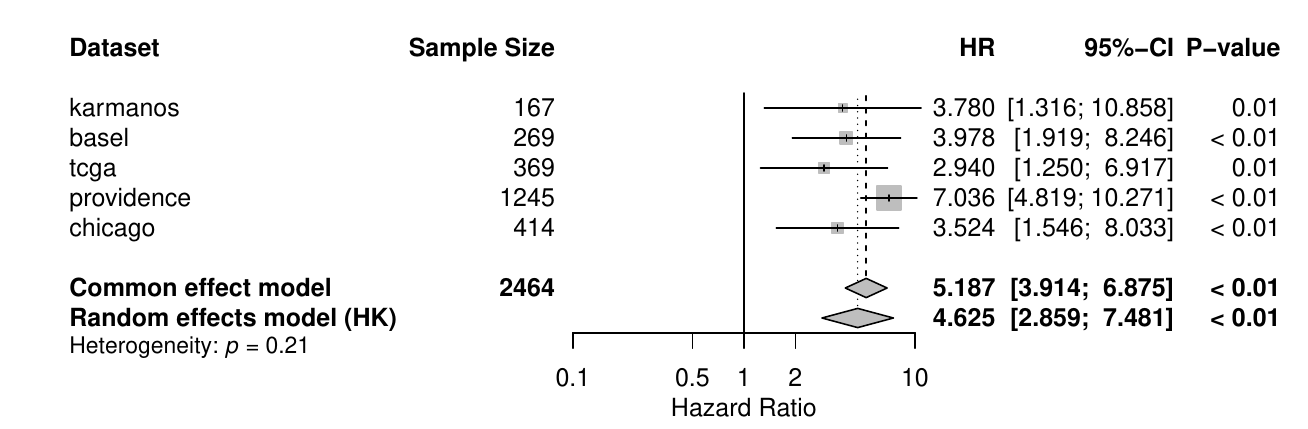}
    \end{subfigure}
    \begin{subfigure}{0.45\textwidth}
        \caption{HR+ HER2\textminus\ ET only}
        \centering
        \includegraphics[trim={180 180 200 200}, clip, width=\linewidth]{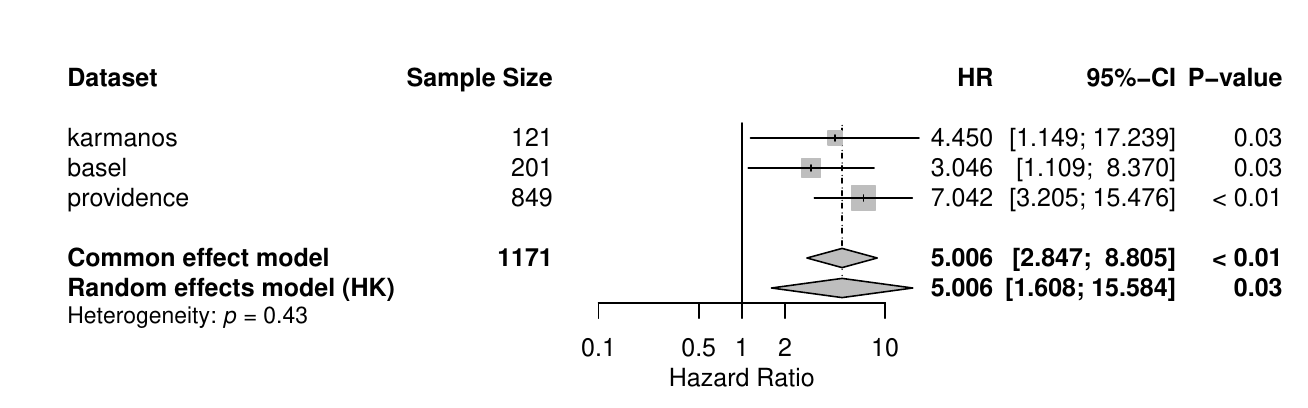}
    \end{subfigure}
    \begin{subfigure}{0.45\textwidth}
        \caption{HR+ HER2\textminus\ ET+CT}
        \centering
        \includegraphics[trim={180 180 200 200}, clip, width=\linewidth]{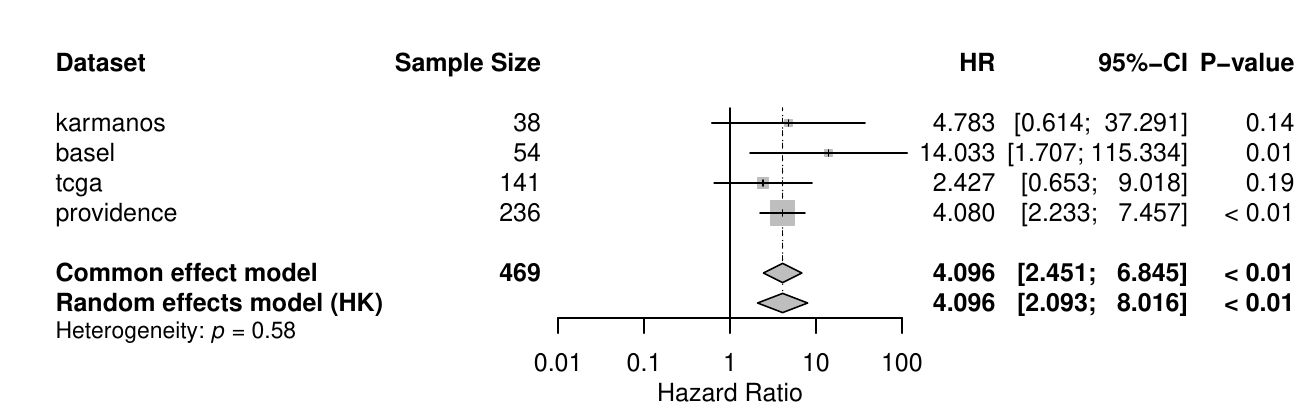}
    \end{subfigure}
    \caption{Random effect models for aggregating hazard ratios based on continuous AI score for different patient subgroups.}
    \label{fig:mixed-effect-subgroups-dfi-ai}
\end{figure}

Here, we present complete results different subgroups for the primary endpoint of DFI. C-index is shown in Figure \ref{fig:mixed-effect-subgroups-dfi-cindex} and hazard ratios is shown in Figure \ref{fig:mixed-effect-subgroups-dfi-ai}. These results include our model’s performance across various endpoints, the performance of the model on individual datasets, along with pooled outcomes derived from a random effects model.

\subsection{Forest plots for the independent performance of Oncotype DX and the AI test}
\label{appendix:forest-plots-oncotype-comparison}

\begin{figure}[htbp]
    \centering
    \begin{subfigure}{0.45\textwidth}
        \caption{Oncotype Score (Hazard Ratio)}
        \centering
        \includegraphics[trim={180 180 200 200}, clip, width=\linewidth]{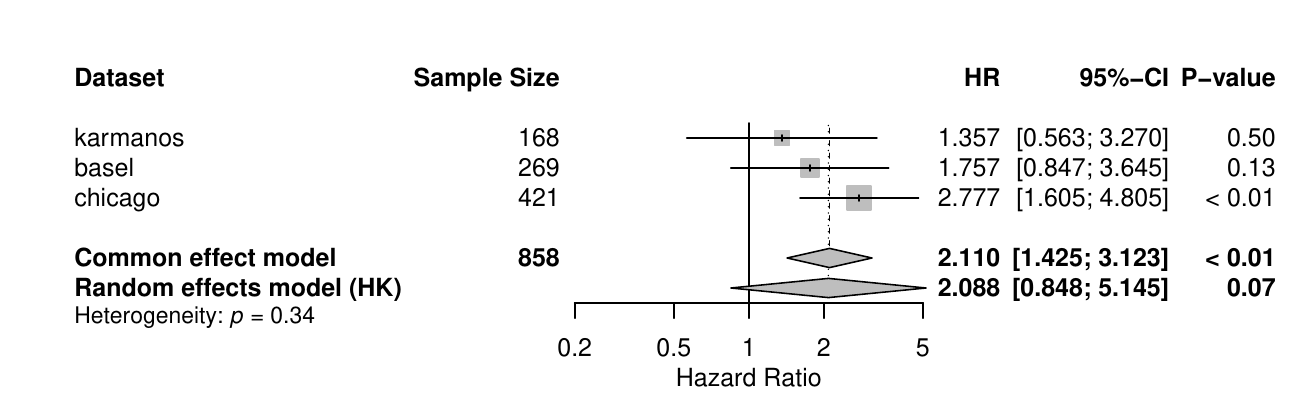}
    \end{subfigure}
    \begin{subfigure}{0.45\textwidth}
        \caption{Multi-modal AI (Hazard Ratio)}
        \centering
        \includegraphics[trim={180 180 200 200}, clip, width=\linewidth]{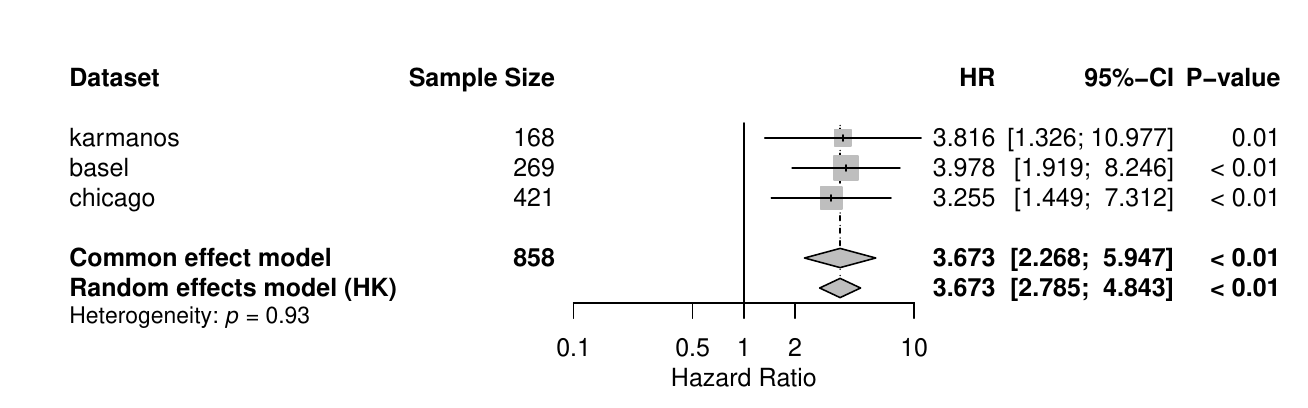}
    \end{subfigure}
    \begin{subfigure}{0.45\textwidth}
        \caption{Oncotype Score (C-index)}
        \centering
        \includegraphics[trim={180 180 270 220}, clip, width=\linewidth]{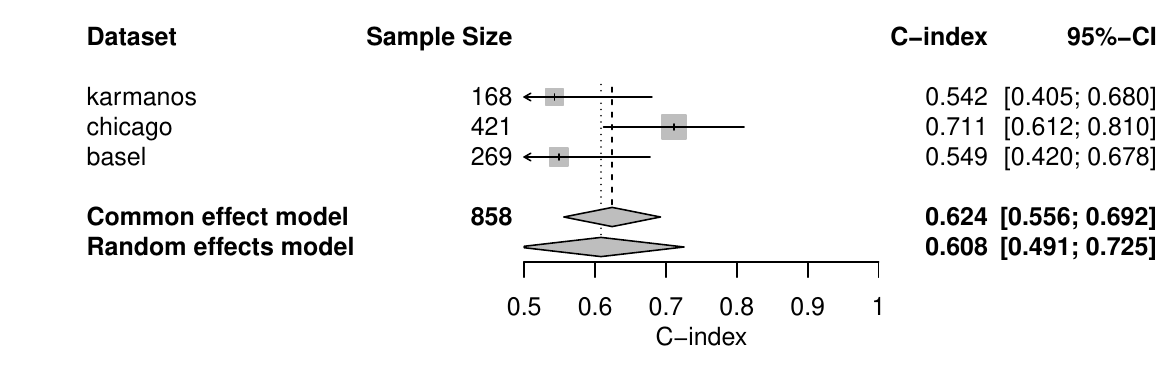}
    \end{subfigure}
    \begin{subfigure}{0.45\textwidth}
        \caption{Multi-modal AI (C-index)}
        \centering
        \includegraphics[trim={180 180 270 220}, clip, width=\linewidth]{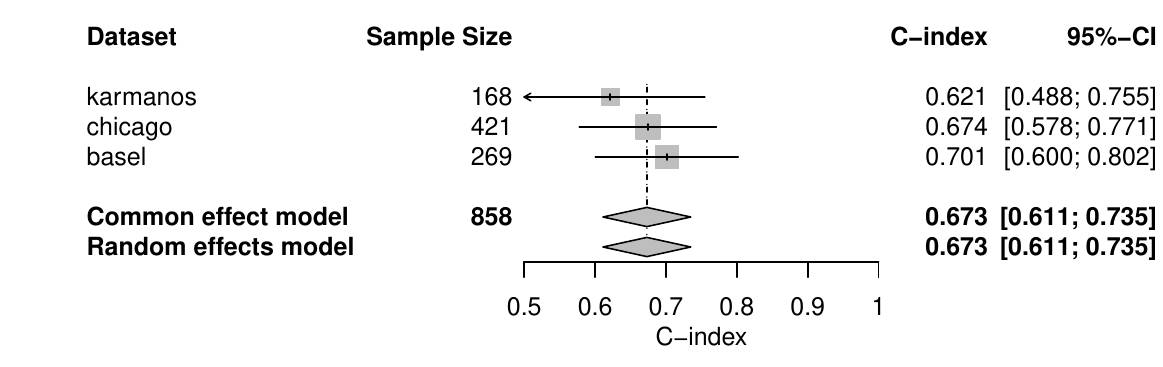}
    \end{subfigure}
    \caption{Random effect models for comparing the hazard ratios between Oncotype and the AI test for the primary DFI endpoint.}
    \label{fig:mixed-effect-subgroups-dfi-oncotype}
\end{figure}

Full results from the random effects model comparing Oncotype and AI test performance, evaluated in patients with Oncotype scores, are presented in Figure \ref{fig:mixed-effect-subgroups-dfi-oncotype}

\end{document}

%% file: figures/plots_for_paper_updated20241027/cox_multivariate_pooled_datasets/clinical_ensemble_x5_pathology_ensemble_x5_dataset_indicator.tex
\begin{tabular}{lll}
\toprule
{} &              All patients &              Oncotype only \\
\midrule
Events              &                       279 &                         67 \\
Total               &                      3502 &                        858 \\
Clinical            &  1.86 (1.67-2.07, p=0.00) &   1.58 (1.21-2.06, p=0.00) \\
Pathology           &  2.35 (1.51-3.66, p=0.00) &  8.82 (3.30-23.53, p=0.00) \\
Chicago dataset     &  0.48 (0.27-0.84, p=0.01) &   0.34 (0.18-0.62, p=0.00) \\
Karmanos dataset    &  1.12 (0.60-2.08, p=0.73) &   0.89 (0.47-1.70, p=0.72) \\
Nightingale dataset &  0.67 (0.44-1.04, p=0.07) &                          - \\
TCGA dataset        &  0.50 (0.30-0.84, p=0.01) &                          - \\
\bottomrule
\end{tabular}

%% file: figures/plots_for_paper_updated20241027/cox_multivariate_pooled_datasets/combined_models.tex
\begin{tabular}{llll}
\toprule
{} &              All patients &    Patients with Oncotype & Oncotype intermediate risk \\
\midrule
Events              &                       279 &                        67 &                         39 \\
Total               &                      3502 &                       858 &                        526 \\
AI test             &  3.63 (2.99-4.40, p=0.00) &  3.46 (2.16-5.54, p=0.00) &   3.45 (1.85-6.42, p=0.00) \\
Chicago dataset     &  0.49 (0.28-0.87, p=0.01) &  0.43 (0.24-0.77, p=0.00) &   0.41 (0.19-0.90, p=0.03) \\
Karmanos dataset    &  1.16 (0.63-2.15, p=0.63) &  1.19 (0.64-2.22, p=0.58) &   1.65 (0.76-3.57, p=0.21) \\
Nightingale dataset &  0.69 (0.45-1.06, p=0.09) &                         - &                          - \\
TCGA dataset        &  0.55 (0.34-0.89, p=0.01) &                         - &                          - \\
\bottomrule
\end{tabular}

%% file: figures/plots_for_paper_updated20241027/cox_multivariate_pooled_datasets/grade_race_age_oncotype_ai_test.tex
\begin{tabular}{lll}
\toprule
{} &              without AI test &           with AI test \\
\midrule
AI Test             &                         - &  3.11 (1.91-5.09, p=0.00) \\
Grade               &  1.38 (0.89-2.14, p=0.15) &  1.24 (0.80-1.94, p=0.34) \\
Race: Other/Unknown &  0.64 (0.21-1.94, p=0.43) &  0.72 (0.24-2.16, p=0.56) \\
Race: Asian         &  0.43 (0.06-3.22, p=0.41) &  0.37 (0.05-2.77, p=0.33) \\
Race: Black         &  1.15 (0.57-2.32, p=0.70) &  1.22 (0.61-2.45, p=0.58) \\
Oncotype            &  1.76 (1.12-2.77, p=0.01) &  1.47 (0.93-2.30, p=0.10) \\
Chicago dataset     &  0.24 (0.07-0.80, p=0.02) &  0.30 (0.09-1.00, p=0.05) \\
Karmanos dataset    &  0.81 (0.31-2.14, p=0.67) &  0.91 (0.35-2.41, p=0.85) \\
\bottomrule
\end{tabular}